\documentclass[%
 aip,
rsi,%
 amsmath,amssymb,
reprint,%
]{revtex4-1}
\usepackage{subcaption}
\usepackage{courier}
\usepackage{color}
\usepackage{listings}
\usepackage{soul}
\usepackage{graphicx}
\usepackage{dcolumn}
\usepackage{bm}
\usepackage{caption}
\usepackage{subcaption}
\usepackage{xfrac}
\pdfoutput=1

\begin{document}

\preprint{AIP/123-QED}

\title{Rapid Time Series Prediction with a Hardware-Based Reservoir Computer}

\author{Daniel Canaday, Aaron Griffith, and Daniel J. Gauthier}
\affiliation{ 
Department of Physics, Ohio State University, 191 West Woodruff Ave., Columbus, Ohio 43210 USA
}%

\date{\today}

\begin{abstract}
Reservoir computing is a neural network approach for processing time-dependent signals that has seen rapid development in recent years. Physical implementations of the technique using optical reservoirs have demonstrated remarkable accuracy and processing speed at benchmark tasks. However, these approaches require an electronic output layer to maintain high performance, which limits their use in tasks such as time-series prediction, where the output is fed back into the reservoir. We present here a reservoir computing scheme that has rapid processing speed both by the reservoir and the output layer. The reservoir is realized by an autonomous, time-delay, Boolean network configured on a field-programmable gate array. We investigate the dynamical properties of the network and observe the fading memory property that is critical for successful reservoir computing. We demonstrate the utility of the technique by training a reservoir to learn the short- and long-term behavior of a chaotic system. We find accuracy comparable to state-of-the-art software approaches of similar network size, but with a superior real-time prediction rate up to 160 MHz.
\end{abstract}

\maketitle

\begin{quotation}
Reservoir computers are well-suited for machine learning tasks that involve processing time-varying signals such as those generated by human speech, communication systems, chaotic systems, weather systems, and autonomous vehicles. Compared to other neural network techniques, reservoir computers can be trained using less data and in much less time. They also possess a large network component, called the reservoir, that can be re-used for different tasks. These advantages have motivated searches for physical implementations of reservoir computers that achieve high-speed and real-time information processing, including opto-electronic and electronic devices.  Here, we develop an electronic approach using an autonomous, time-delay, Boolean network configured on a field-programmable gate array (FPGA). These devices allow for complex networks consisting of 1,000's of nodes with arbitrary network topology. Time-delays can be incorporated along network links, thereby allowing for extremely high-dimension reservoirs. The characteristic time scale of a network node is less than a nanosecond, allowing for information processing in the GHz regime. Further, because the reservoir state is Boolean rather than real-valued, calculation of an output from the reservoir state can be done rapidly with synchronous FPGA logic. We use such a reservoir computer for the challenging task of forecasting the dynamics of a chaotic system. This work paves the way for low-cost, compact reservoir computers that can be embedded in various commercial and industrial systems for real-time information processing. 
\end{quotation}

\section{Introduction}

	There is considerable interest in the machine learning community in using recurrent neural networks (RNN) for processing time-dependent signals.\cite{zhang2014comprehensive,lukovsevivcius2009reservoir,gauthier2018reservoir} Under some mild assumptions, these types of networks are universal approximators of dynamical systems,\cite{funahashi1993approximation} similarly to how multilayer feedforward neural networks are universal approximators of static maps.\cite{hornik1989multilayer} Many machine learning and artificial intelligence tasks, such as dynamical system modeling, human speech recognition, and natural language processing are intrinsically time-dependent tasks, and thus are more naturally handled within a time-dependent, neural-network framework.
    
	Though they have high expressive power, RNNs are difficult to train using gradient-descent-based methods.\cite{hochreiter2001gradient} One approach to efficiently and rapidly train an RNN is known as reservoir computing (RC). In RC, the network is divided into input nodes, a bulk collection of nodes known as the \textit{reservoir}, and output nodes, such that the only recurrent links are between reservoir nodes. Training involves only adjusting the weights along links connecting the reservoir to the output nodes and not the recurrent links in the reservoir. This approach displays state-of-the-art performance in a variety of time-dependent tasks, including chaotic time series prediction,\cite{li2012} system identification and control,\cite{Jaeger2003} and spoken word recognition,\cite{fernando2003} all with remarkably short training times in comparison to other neural-network approaches.
        
	Recently, implementations of reservoir computing using dedicated hardware have achieved much attention, particularly those based on delay-coupled photonic systems.\cite{larger2012photonic,vandoorne2008toward,nakayama2016laser} These devices allow for reservoir computing at extremely high speeds, including the classification of spoken words at a rate of millions of words per second.\cite{larger2017high} There is also the potential to form the input and output layers out of optics as well, resulting in an all-optical computational device.\cite{bueno2017conditions,duport2012all} However, these devices are not well-equipped to handle tasks such as time-series prediction, which require the input and output layer to be coupled.
    
    Here, we present a hardware implementation of RC based on an autonomous, time-delay, Boolean network realized on a readily-available platform known as a field-programmable gate array (FPGA). This approach allows for a seamless coupling of reservoir to output due to the spatially simple nature of the reservoir state and the fact that matrix multiplication can be realized with compact Boolean logic. Together with the parallel nature of the network, this allows for up to 10 times faster information processing than delay-coupled photonic devices.\cite{larger2017high} We apply our implementation to the challenging task of predicting the behavior of a chaotic dynamical system. We find prediction accuracy similar to software-based techniques of similar network size and achieve a record-high real-time prediction rate.\cite{antonik2016}

	The rest of this article is organized as follows: we describe the RC technique in general terms, detailing the necessary components and their features in Sec. II; we discuss our approach to realizing these features in an efficient manner on an FPGA in Sec. III-V; we discuss the performance of our approach to prediction of the Mackey-Glass system in Sec. VI-VII; and we conclude with a discussion of our results in Sec. VIII.
    
\section{Reservoir Computing for Time-Series Prediction}
	Reservoir computing is a concept introduced independently by Jaeger\cite{jaeger2001} and Maass\cite{maass2002} under the names Echo State Network (ESN) and Liquid State Machine (LSM), respectively. In Jaeger's technique, a network of recurrently connected sigmoidal nodes (the reservoir) with state $\textbf{X}(t)$ is excited by a time-dependent input signal $\textbf{u}(t)$. The reservoir is observed during some training period, an approximate linear transformation from $\textbf{X}(t)$ to a desired signal $\textbf{v}_d(t)$ is identified via linear regression, and this linear transformation forms the readout layer. These signals and their relations to one another are illustrated in Fig. 1a. The LSM technique has the same features, but uses a pool of spiking nodes to form the reservoir.

	The two approaches described by Jaeger and Maass are apparently similar, and indeed are two particular implementations of RC. As a class of techniques, RC can be defined quite broadly, and we do so as follows. Given an input signal $\textbf{u}(t)$ and a desired output signal $\textbf{v}_d(t)$, a reservoir computer constructs a mapping from $\textbf{u}(t)$ to $\textbf{v}_d(t)$ with the following steps:

\begin{itemize}
\item create a randomly parameterized network of nodes and recurrent links called the \textit{reservoir} with state $\textbf{X}(t)$ and dynamics described by $\dot{\textbf{X}}(t) = \textbf{f}\big(\textbf{X}(t), \textbf{u}\left(t\right)\big)$;
\item excite the reservoir with an input signal $\textbf{u}(t)$ over some training period and observe the response of the reservoir;
\item form a readout layer that transforms the reservoir state $\textbf{X}(t)$ to an output $\textbf{v}(t)$, such that $\textbf{v}(t)$ well approximates $\textbf{v}_d(t)$ during the training period.
\end{itemize}
Figure 1a contains a schematic representation of the resulting system, which consists of the reservoir, the input signal, the trained readout layer, and output signal. Note that we make no assumptions about the dynamics $\textbf{f}$. In general, it may include discontinuities, time-delays, or have components simply equal to $\textbf{u}(t)$ (\textit{i.e.}, the reservoir may include a direct connection from input to output).

	Reservoir computing demonstrates remarkable success at predicting a chaotic time series, among other applications. The goal of this task is to predict the output of an unknown dynamical system after a training period. In the context of RC, this is accomplished by setting $\textbf{v}_d(t) = \textbf{u}(t)$, \textit{i.e.}, by training the reservoir computer to reproduce its inputs. Then, after training is complete, we replace $\textbf{u}(t)$ with $\textbf{v}(t)$ and allow the newly-formed autonomous system to evolve in time beyond the end of the training period. This closed-loop system is illustrated in Fig. 1b and consists of the same components as in the general picture, but the input and output are the same signal. This scheme can predict accurately the short-term behavior of a variety of systems, including the Mackey-Glass,\cite{jaeger2001} Lorenz,\cite{pathak2017} and Kuramoto-Sivashinsky spatial-temporal\cite{pathak2017} systems using a software simulation of the reservoir. A reservoir computer trained in this manner can also learn the long-term behavior of complex systems, generating the true attractor of the target system and replicating its Lyapunov spectrum.\cite{pathak2017}

\begin{figure}[h]
\centering
\begin{subfigure}[b]{0.5\textwidth}
\captionsetup{font=sf}
\centering
\caption{}
\includegraphics[width=\textwidth]{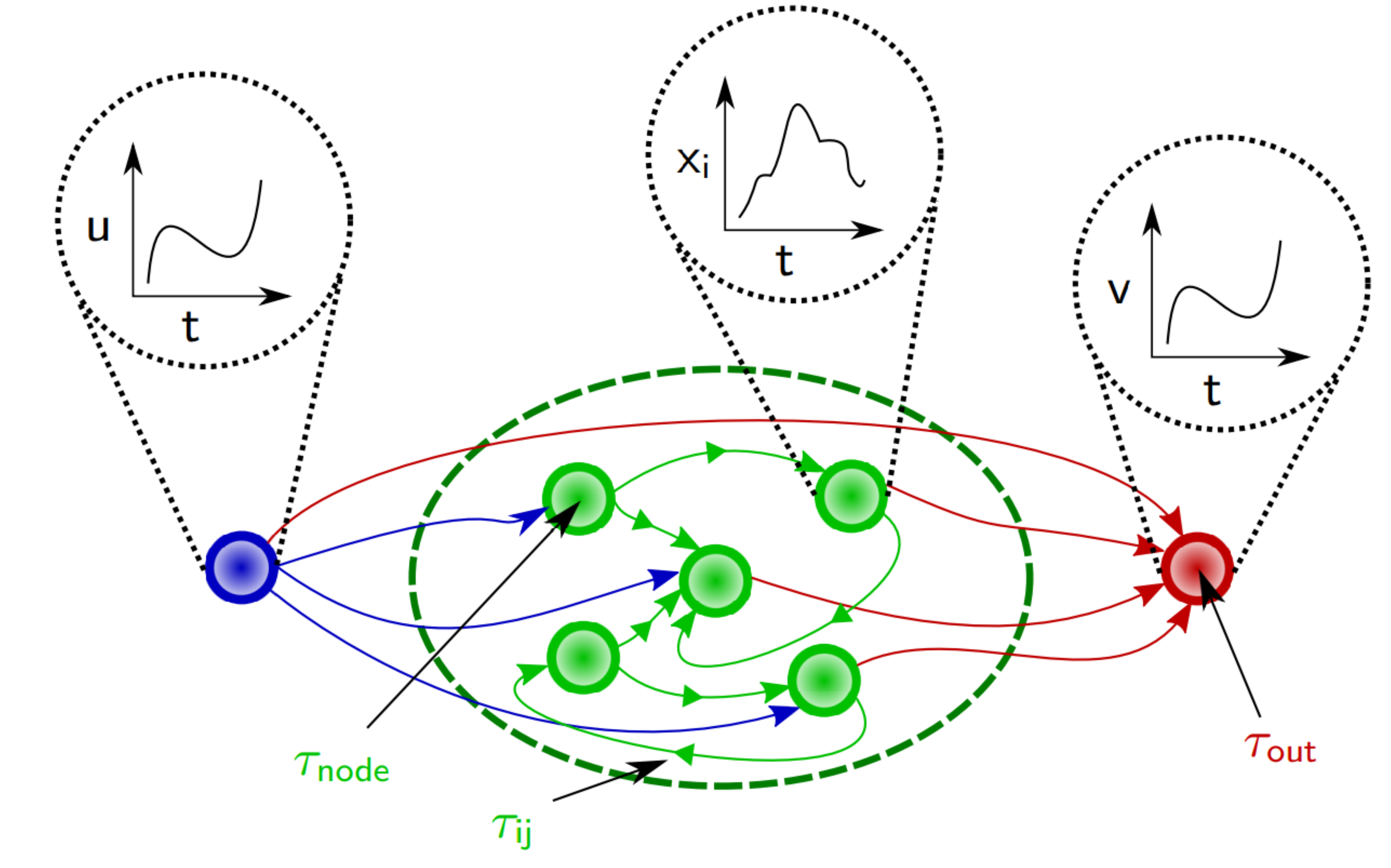}
\end{subfigure}
\centering
\begin{subfigure}[b]{0.5\textwidth}
\captionsetup{font=sf}
\centering
\caption{}
\includegraphics[width=\textwidth]{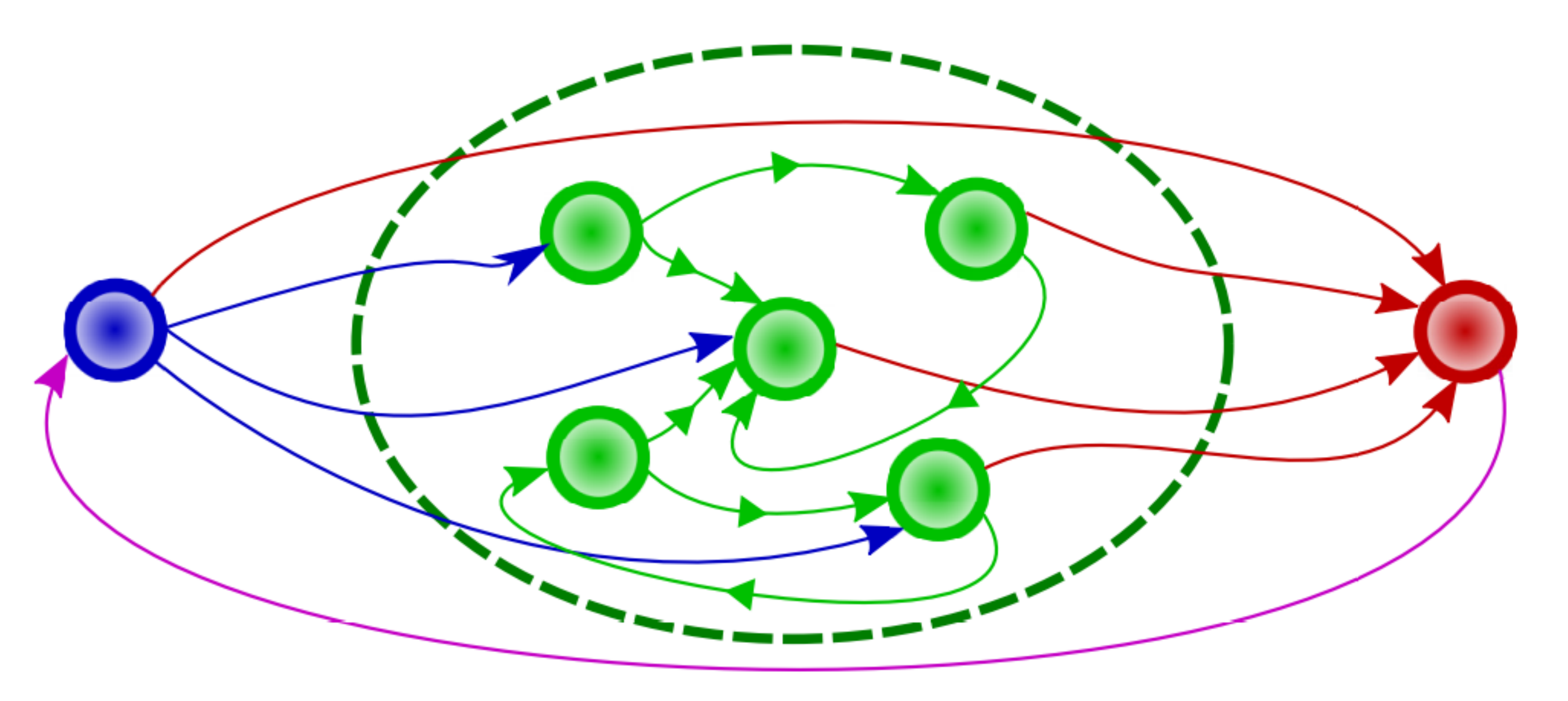}
\end{subfigure}
\caption{Schematic representation of the RC scheme. (a) A general reservoir computer learns to map an input onto a desired output. The network dynamics may contain propagation delays along the links (denoted by $\tau_{ij}$) or through nodes (such as through the output layer, denoted by $\tau_{out}$). (b) For the particular task of predicting a signal, the reservoir is trained so that the target output is equal to the input. After training, the output is fed back into the reservoir, resulting in an autonomous dynamical system. If properly trained, the autonomous reservoir serves as a model for the dynamics that generated the input signal.}
\end{figure}

	Although training the network consists only of identifying optimal parameters in the readout layer, there are a variety of factors in designing the reservoir that impact the success of the scheme. These factors include:

\textbf{Matching time scales}. In general, both the reservoir and the source of the signal $\textbf{u}(t)$ are dynamical systems with their own characteristic time scales. These time scales must be similar for the reservoir to produce $\textbf{v}(t)$.\cite{schrauwen2007introduction} For software based approaches to RC, these scales are matched by tuning the reservoir's temporal properties through accessible reservoir parameters, such as the response time $\tau_{node}$ of reservoir nodes. However, with hardware-based approaches, the parameters controlling the time-scale of reservoir dynamics are often more rigid. We compensate for this by adjusting the time scale of the input signal (see Sec. IV) and adding delays to the links within the reservoir (see Sec. IIIa). 

\textbf{Reservoir Memory}. It is generally believed, as was observed by Jaeger and Maass in their respective architectures and as has been explored more generally,\cite{wyffels2010} that a good reservoir for RC is a system that possesses \textit{fading memory}. That is, the reservoir state contains information about the input signal $\textbf{u}(t)$, but the effect of small differences in $\textbf{u}(t)$ dissipate over time. This is often referred to as the echo-state property in the context of ESNs and is described in greater detail in Sec. IIIb. We find that the autonomous reservoirs considered here have the fading memory property. Further, we find that the characteristic time scale over which small differences dissipate can be tuned by adding delays to the links within the reservoir.

\textbf{Coupling to Input}. Each RC implementation couples $\textbf{u}(t)$ to the reservoir in a very technique-dependent way, such as spike-encoding in LSMs or by consideration of so-called ``virtual nodes" in photonic reservoirs.\cite{appeltant2011} The coupling in our FPGA-based approach is complicated by the fact that nodes execute Boolean functions, whereas the input signal $\textbf{u}(t)$ is a large-bit representation of a real number. We must also consider, as with most techniques for processing physical data, the limited precision and sampling rate of the input signal. The sampling rate is particularly relevant for our physical reservoir computer, as the the reservoir nodes have their own, fixed characteristic time scale. These issues are discussed in Sec. IV.

\textbf{Calculating $\textbf{v}(t)$}. In software-based reservoir computing schemes, the readout layer performs its operation effectively instantaneously as far as the simulation is concerned. However, this is not possible when the reservoir is a continuously-evolving physical system. There is a finite time required to calculate $\textbf{v}(t)$, which can be interpreted as a propagation delay $\tau_{out}$ (see Fig. 1a) through the readout layer and ultimately limits the rate at which predictions can be made in closed-loop operation. Consequently, $\textbf{v}(t)$ must be calculated from a measurement of $\textbf{X}(t-\tau_{out})$ for the predicted output to be ready to be fed back into the input at time $t$.

	The goal of this work is to demonstrate a technique for realizing a hardware implementation of RC with minimal output delay so that predictions can be made as rapidly as possible. In the next few sections, we detail the construction of the various components of the reservoir computer illustrated in Fig. 1 and how they address the general RC properties outlined in this section.     

\section{Autonomous Boolean Reservoir}
    
   We propose a reservoir construction based on an autonomous, time-delay, Boolean reservoir realized on an FPGA. By forming the nodes of the reservoir out of FPGA elements themselves, this approach exhibits faster computation than FPGA-accelerated neural networks,\cite{schrauwen2008compact, alomar2016fpga} which require explicit multiplication, addition, and non-linear transformation calculations at each time-step. Our approach also has the advantage of realizing the reservoir and the readout layer on the same platform without delays associated with transferring data between different hardware. Finally, due to the Boolean-valued state of the reservoir, a linear readout layer $\big[\textbf{v}(t) = \textbf{W}_{out}\textbf{X}(t)\big]$ is reduced to an addition of real numbers rather than a full matrix multiplication. This allows for much shorter total calculation time and thus faster real-time prediction than in opto-electronic RC.\cite{antonik2016}

	Our choice of reservoir is further motivated by the observation that Boolean networks with time-delay can exhibit complex dynamics, including chaos.\cite{zhang2009} In fact, a single XOR node with delayed feedback can exhibit a fading memory condition and is suitable for RC on simple-tasks such as binary pattern recognition.\cite{Haynes2015} 
    
    It has been proposed\cite{apostel2017} that individual FPGA nodes have dynamics that can be described by the Glass model\cite{glass1973} given by
    
\begin{equation}
\gamma_i \dot{x}_i = -x_i + \Lambda_i(X_{i1}, X_{i2}, ...),
\end{equation}
\begin{equation}
X_i = \begin{cases}
1 \textit{ if } x_i \geq q_i, \\
0 \textit{ if } x_i < q_i, \\
\end{cases}
\end{equation}
where $x_i$ is the continuous variable describing the state of the node, $\gamma_i$ describes the time-scale of the node, $q_i$ is a thresholding variable, and $\Lambda_i$ is the Boolean function assigned to the node. The thresholded Boolean variable $X_{ij}$ is the $j_{th}$ input to the $i_{th}$ node.

	We construct our Boolean reservoir by forming networks of nodes described by Eq. 1-2 and the Boolean function
    
\begin{equation}
\Lambda_i = \Theta \left( \sum_j W^{ij} X_j + W_{in}^{ij} u_j \right),
\end{equation}
where $u_j$ are the bits of the input vector $\textbf{u}$, $\textbf{W}$ is the reservoir-reservoir connection matrix, $\textbf{W}_{in}$ is the input-reservoir connection matrix, and $\Theta$ is the Heaviside step function defined by 

\begin{equation}
\Theta(x) = 
\begin{cases}
1 \textit{ if } x > 0, \\
0 \textit{ if } x \leq 0.
\end{cases}
\end{equation}

	The matrices $\textbf{W}$ and $\textbf{W}_{in}$ are chosen as follows. Each node receives input from exactly $k$ other randomly chosen nodes, thus determining $k$ non-zero elements of each row of $\textbf{W}$. The non-zero elements of $\textbf{W}$ are given a random value from a uniform distribution between $-1$ and $1$. The maximum absolute eigenvalue (spectral radius) of the matrix $\textbf{W}$ is calculated and used to scale $\textbf{W}$ such that its spectral radius is $\rho$. A proportion $\sigma$ of the nodes are chosen to receive input, thus determining the number of non-zero rows of $\textbf{W}_{in}$. The non-zero values of $\textbf{W}_{in}$ must be chosen carefully (see Sec. IV), but we note here that the scale of $\textbf{W}_{in}$ does not need to be tuned, as it is apparent from Eq. 3 that only the relative scale of $\textbf{W}$ and $\textbf{W}_{in}$ determines $\Lambda_i$. 

The three parameters defined above--$k, \rho$, and $\sigma$--are the three hyperparameters that characterize the topology of the reservoir. We introduce a final parameter $\bar{\tau}$ in the next section, which characterizes delays introduced along links between nodes. Together, these four hyperparameters describe the reservoirs that we investigate in this work.

\subsection{Matching Time Scales with Delays}

	The presence of the $-x_i$ term in Eq. 1 represents the sluggish response of the node, \textit{i.e.}, its inability to change its state instantaneously. This results in an effective propagation delay of a signal through the node. We can take advantage of this phenomenon by connecting chains of pairs of inverter gates between nodes. These inverter gates have dynamics described by Eq. 1-2 and
    
\begin{equation}
\Lambda_i(X) = 
\begin{cases}
0 \textit{ if } X = 1,\\
1 \textit{ if } X = 0,
\end{cases}
\end{equation}
Note that the propagation delay through these nodes depends both on $\gamma_i$ and $q_i$, both of which are heterogeneous throughout the chip due to small manufacturing differences. We denote the mean propagation delay through the inverter gates by $\tau_{inv}$, which we measure by recording the oscillation frequencies of variously sized loops of these gates. For the Arria 10 devices considered here,\footnote{We use an Arria 10 SX 10AS066H3F34I2SG chip for the results discussed in this paper.} we find $\tau_{inv} = 0.19 \pm 0.05$ ns.

	We exploit the propagation delays by inserting chains of pairs of inverter gates in between reservoir nodes, thus creating a time-delayed network. We fix the mean delay $\bar{\tau}$ and randomly choose a delay time for each network link. This is similar to how the network topology is chosen by fixing certain hyperparameters and randomly choosing $\textbf{W}$ and $\textbf{W}_{in}$ subject to these parameters. The random delays are chosen from a uniform distribution between $\bar{\tau}/2$ and $3 \bar{\tau}/2$ so that delays on the order of $\tau_{node}$ are avoided. 

    The addition of these delay chains is necessary because the time-scale of individual nodes is must faster than the speed at which synchronous FPGA logic can change the value of the input signal (see Sec. IV). Without any delays, it is impossible to match the time-scales of the input signal with the reservoir state, and we have poor RC performance. We find that the time-scales associated with the reservoir's fading memory are controlled by $\bar{\tau}$, as described in the next section, thus demonstrating that we can tune the reservoir's time-scales with delay lines. 
    
\subsection{Fading Memory}

	For the reservoir to learn about its input sequence, it is believed that it must possess the fading memory property (although more may be required for replicating long-term behavior\cite{lu2018}). Intuitively, this property implies that the reservoir state $\textbf{X}(t)$ is a function of its input history, but is more strongly correlated with more recent inputs. More precisely, the fading memory property states that every reservoir state $\textbf{X}(t_0)$ is uniquely determined by a left-infinite input sequence $\{\textbf{u}(t): t<t_0\}$.
    
    The fading memory property is equivalent \cite{jaeger2001} to the statement that, for any two reservoir states $\textbf{X}_1(t_0)$ and $\textbf{X}_2(t_0)$ and input signal $\{\textbf{u}(t): t>t_0\}$, we have
    
\begin{equation}
\lim_{t\rightarrow \infty} ||\textbf{X}_1(t)-\textbf{X}_2(t)||_2 = 0.
\end{equation}
Also of interest is the characteristic time-scale over which this limit approaches zero, which may be understood as the Lyapunov exponent of the coupled reservoir-input system conditioned on the input.
    
    We observe the fading memory property and measure the corresponding time-scale with the following procedure. We prepare two input sequences $\{\textbf{u}_1(i\Delta t);-N\leq i \leq N\}$ and $\{\textbf{u}_2(i\Delta t);-N\leq i \leq N\}$, where $\Delta t$ is the input sample rate (see Sec. IV) and $N$ is an integer such that $N\Delta t$ is sufficiently large. Each $\textbf{u}_1(i\Delta t)$ is drawn from a random, uniform distribution between $-1$ and $1$. For $i\geq 0$, $\textbf{u}_2(i\Delta t) = \textbf{u}_1(i\Delta t)$. For $i<0$, $\textbf{u}_2(i\Delta t)$ is drawn from a random, uniform distribution between $-1$ and $1$. We drive the reservoir with the first input sequence and observe the reservoir response $\{\textbf{X}_1(i\Delta t); -N\leq i \leq N\}$. After the reservoir is allowed to settle to its equilibrium state, we drive it with the second input sequence and observe $\{\textbf{X}_2(i\Delta t); -N\leq i \leq N\}$. The reservoir is perturbed to effectively random reservoir states $\textbf{X}_1(0)$ and $\textbf{X}_2(0)$, because the input sequences are unequal for $i<0$. For $i\geq 0$, the input sequences are equal, and the difference in Eq. 6 can be calculated. 
    
    For a given reservoir, this procedure is repeated 100 times with different input sequences. For each pair of sequences, the state difference is fit to exp$(-t/ \lambda)$, and the $\lambda$'s are averaged over all 100 sequences. We call $\lambda$ the reservoir's \textit{decay time}. We find $\lambda > 0$ for every reservoir examined, demonstrating the usefulness of the chosen form of $\Lambda_i$ in Eq. 3.
    
    We explore the dependence of the decay time as a function of hyperparameter $\bar{\tau}$. As seen from Fig. 2, the relationship is approximately linear for fixed $k, \rho$, and $\sigma$. This is consistent with $\bar{\tau}$ being the dominate time-scale of the reservoir rather than $\tau_{node}$, which is our motivation for including delay lines in our reservoir construction. The dependence of $\lambda$ on the other hyperparameters defined in Sec. III are explored in Sec. VI along with corresponding results on a time-series prediction task.
    
\begin{figure}[h]
\centering
\begin{subfigure}[b]{0.4\textwidth}
\captionsetup{font=sf}
\centering
\includegraphics[width=\textwidth]{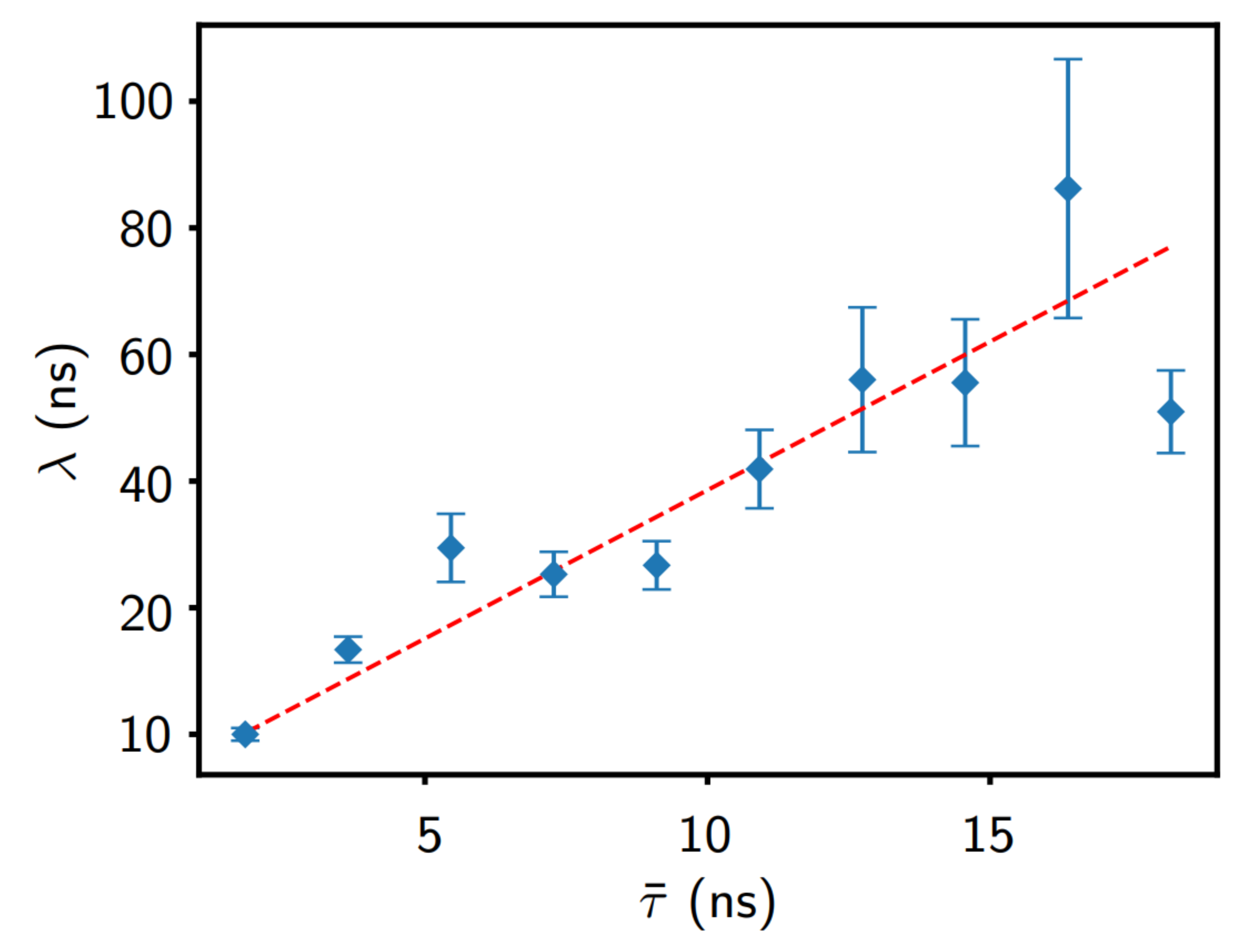}
\end{subfigure}
\caption{Experimental observation of the fading memory property and decay time for varying $\bar{\tau}$. The network has 100 nodes and hyperparameters $k=2, \rho=1.5$, and $\sigma = 0.75$. Statistics are generated by testing five reservoirs for each set of hyperparameters. Vertical error bars represent the standard error of the mean. The relationship is approximately linear with a slope of $3.99 \pm 0.45$.}
\end{figure} 

\section{Input Layer}

	As discussed in Sec. III, our reservoir implementation is an autonomous system without a global clock, allowing for continuously evolving dynamics. However, the input layer is a synchronous FPGA design that sets the state of the input signal $\textbf{u}(t)$. Prior to operation, a sequence of values for $\textbf{u}(t)$ is stored in the FPGA memory blocks. During the training period, the input layer sequentially changes the state of the input signal according to the stored values.

	For the prediction task, the stored values of $\textbf{u}(t)$ are observations of some time-series from $t=-T_{train}$ to $t=0$. This signal maybe defined on the entire real interval $[-T_{train},0]$, but only a finite sampling may be stored in the FPGA memory and presented as input to the reservoir. The signal may also take real values, but only a finite resolution at each sampling interval may be stored. The actual input signal $\textbf{u}(t)$ in Fig. 1 is thus discretized in two ways:
\begin{itemize}
\item $\textbf{u}(t)$ is held constant along intervals of length $t_{sample}$;
\item $\textbf{u}(t)$ is approximated by an $n-$bit representation of real numbers.
\end{itemize}
A visualization of these discretizations is in Fig. 3. Note that $t_{sample}$ is a physical unit of time, whereas $\Delta t$ has whatever units (if any) in which the non-discretized time-series is defined. 

	As pointed out in Sec. III, $t_{sample}$ may be no smaller than the minimum time in which the clocked FPGA logic can change the state of the input signal, which is approximately 5 ns on the Arria 10 device considered here. However, we show in Sec. V that $t_{sample}$ must be greater than or equal to $\tau_{out}$, which generally cannot be made as short as 5 ns. 

\begin{figure}[h]
\centering
\begin{subfigure}[b]{0.23\textwidth}
\captionsetup{font=sf}
\centering
\caption{}
\includegraphics[width=\textwidth]{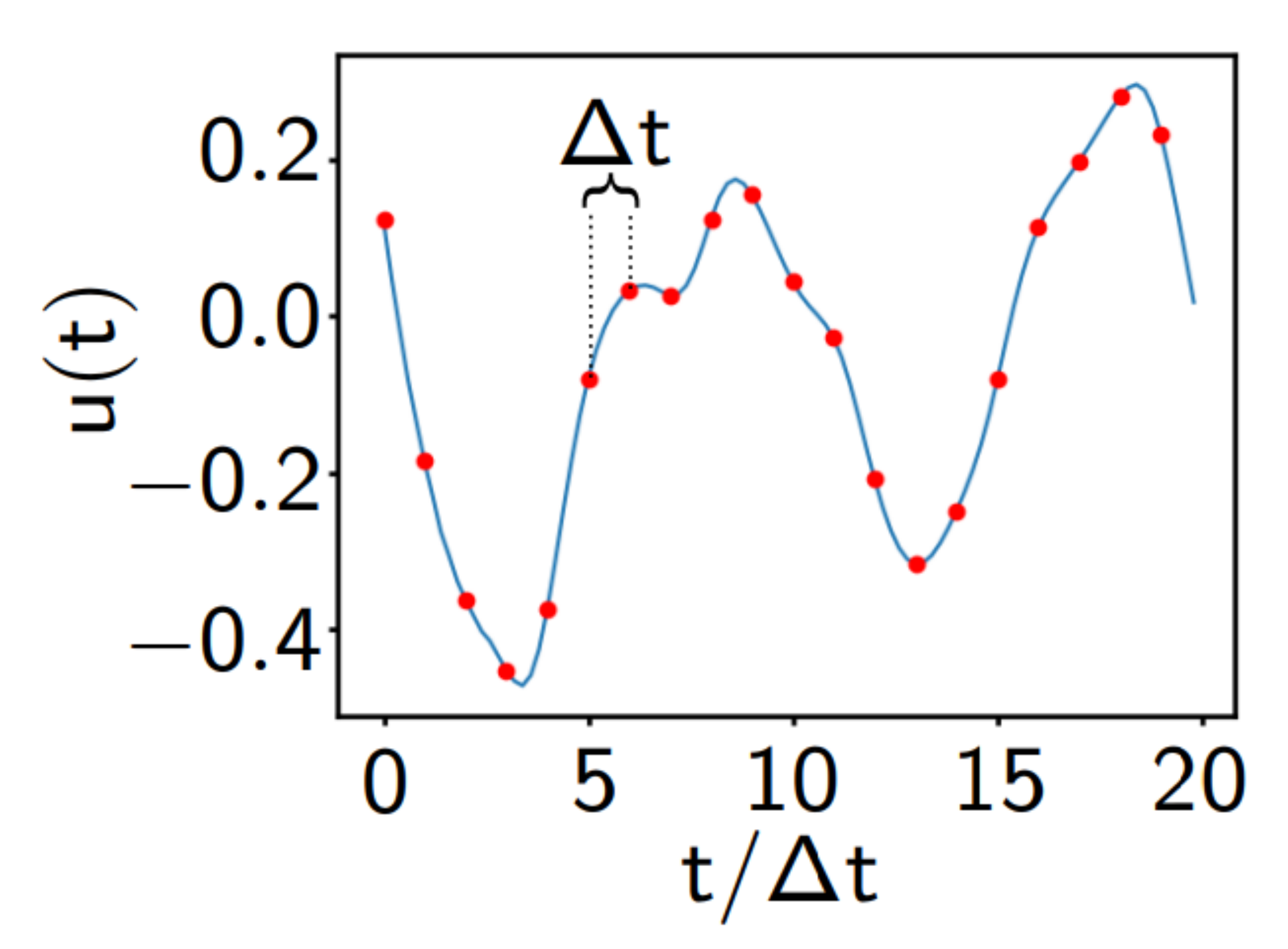}
\end{subfigure}
\centering
\begin{subfigure}[b]{0.23\textwidth}
\captionsetup{font=sf}
\centering
\caption{}
\includegraphics[width=\textwidth]{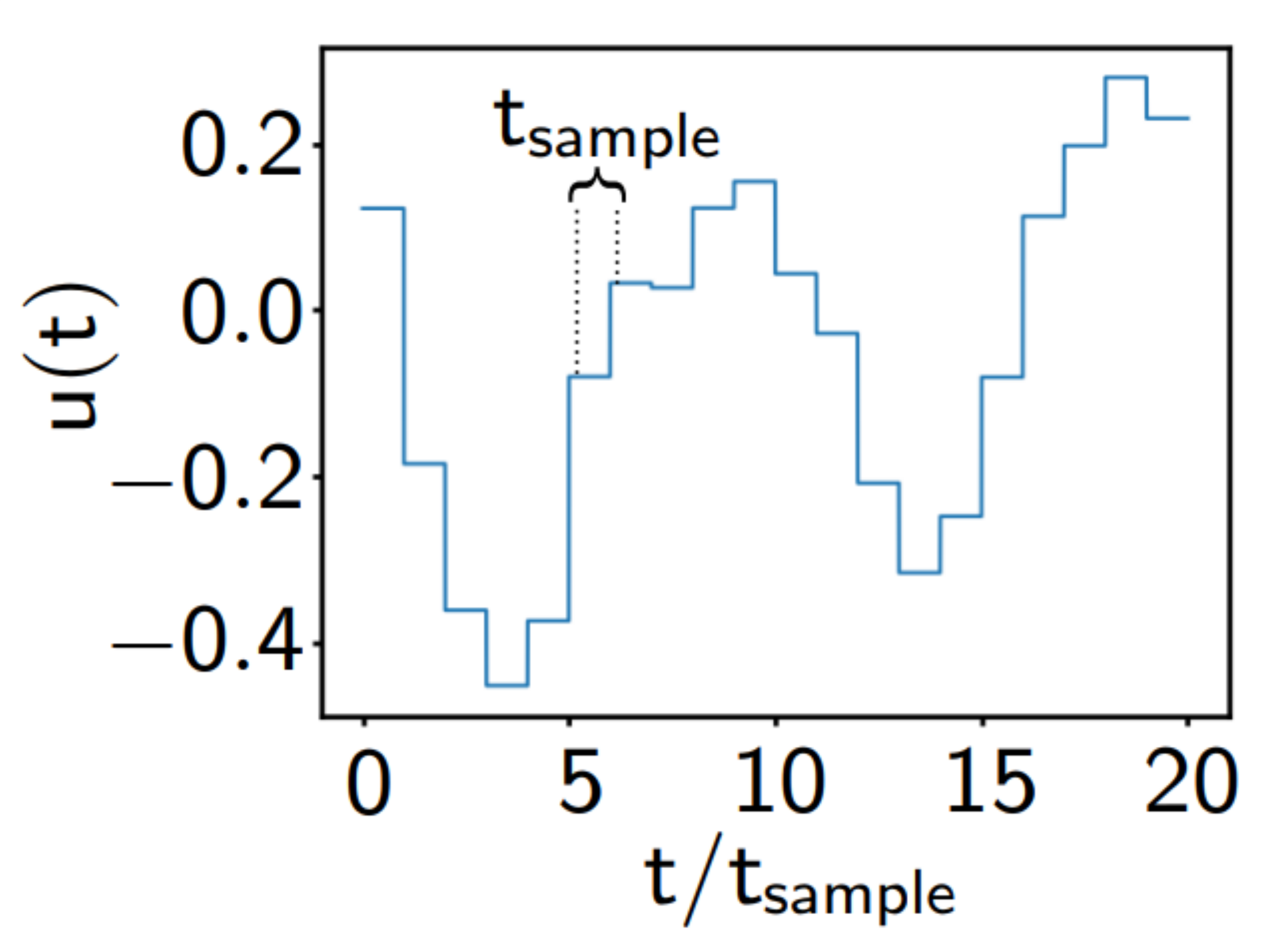}
\end{subfigure}
\caption{A visualization of the discretization of $\textbf{u}(t)$ necessary for hardware computation. (a) In general, the true input signal may be real-valued and defined over a continuous interval. (b) Due to finite precision and sampling time, the actual $\textbf{u}(t)$ seen by the reservoir is held constant over intervals of duration $t_{sample}$ and have finite vertical precision. For the prediction task, $v_d(t) = u(t)$, so the output must be discretized similarly.}
\end{figure}

\subsection{Binary Representations of Real Data}

	The Boolean functions described by Eq. 3-4 are defined according to Boolean values $u_j$, which are the bits in the $n-$bit representation of the input signal. If the elements of $\textbf{W}_{in}$ are drawn randomly from a single distribution, then the reservoir state is as much affected by the least significant bit of $\textbf{u}(t)$ as it is the most significant. This leads to the reservoir state being distracted by small differences in the input signal and fails to produce a working reservoir computer. 
    
    For a scalar input $u(t)$, we can correct for this shortcoming by choosing the rows of $\textbf{W}_{in}$ such that 
\begin{equation}
\sum_j W_{in}^{i,j}u_j \approx \tilde{W}_{in}^i u,
\end{equation}
where $\tilde{\textbf{W}}_{in}$ is an effective input matrix with non-zero values drawn randomly between $1$ and $-1$. The relationship is approximate in the sense that $u$ is a real-number and $u_j$ is a binary representation of that number. For the two's complement representation, this is done by choosing

\begin{equation}
W_{in}^{i,j} = 
\begin{cases}
-2^{\left(n-1\right)}\tilde{W}_{in}^i \textit{ if } j=n,\\
+2^{\left(j-1\right)}\tilde{W}_{in}^i \textit{ else }.
\end{cases}
\end{equation}

	A disadvantage of the proposed scheme is that every bit in the representation of $u$ must go to every node in the reservoir. If a node has $k$ recurrent connections, then it must execute a $n+k$ to $1$ Boolean function, as can be seen from Eq. 3. Boolean functions with more inputs take more FPGA resources to realize in hardware, and it takes more time for a compiler to simplify the function. We find that an $8-$bit representation of $u$ is sufficient for the prediction task considered here while maintaining achievable networks. 

\section{Output Layer}

	Similar to the input layer, the output layer is constructed from synchronous FPGA logic. Its function is to observe the reservoir state and, based on a learned output matrix $\textbf{W}_{out}$, produce the output $\textbf{v}(t)$. As noted in Sec. II, this operation requires a time $\tau_{out}$ that we interpret as a propagation delay through the output layer and requires that $\textbf{v}(t)$ be calculated from $\textbf{X}(t-\tau_{out})$. 
    
    For the time-series prediction task, the desired reservoir output $\textbf{v}_d(t)$ is just $\textbf{u}(t)$. As discussed in the previous section, the input signal is discretized both in time and in precision so that the true state of the input signal is similar to the signal in Fig. 3b. Thus, $\textbf{v}(t)$ must be discretized in the same fashion. Note that, because the reservoir state $\textbf{X}(t)$ is Boolean valued, a linear transformation $\textbf{W}_{out}$ of the reservoir state is equivalent to a partial sum of the weights $\textbf{W}_{out}$, where $W_{out}^i$ is included in the sum only if $X_i(t) = 1$. 

	We find that the inclusion of a direct connection (see Sec. II and Fig. 1) greatly improve prediction performance. Though this involves a multiplication of $8-$bit numbers, it only slightly increases $\tau_{out}$ because this multiplication can be done in parallel with the calculation of the addition of the Boolean reservoir state. 

	With the above considerations in mind, the output layer is constructed as follows: on the rising edge of a global clock with period $t_{global}$, the reservoir state is passed to a register in the output layer. The output layer calculates $\textbf{W}_{out}\textbf{X}$ with synchronous logic and in one clock cycle, where the weights $\textbf{W}_{out}$ are stored in on-board memory blocks. The calculated output $\textbf{v}(t)$ is passed to a register on the edge of the global clock. If $t>0$, \textit{i.e.} if the training period has ended, the input layer passes $\textbf{v}(t)$ to the reservoir rather than the next stored value of $\textbf{u}(t)$. 
    
    For $\textbf{v}(t)$ to have the same discretized form as $\textbf{u}(t)$, we must have the global clock period $t_{global}$ be equal to the input period $t_{sample}$, which means the fastest our reservoir computer can produce predictions is once every max$\{\tau_{out},t_{sample}\}$. While $t_{sample}$ is independent of the size of the reservoir and precision of the input, $\tau_{out}$ in general depends on both. We find that $\tau_{out} = 6.25$ ns is the limiting period for a reservoir of 100 nodes, an 8-bit input precision, and the Arria 10 FPGA considered here. Our reservoir computer is therefore able to make predictions at a rate of 160 MHz, which is currently the fastest prediction rate of any real-time RC to the best of our knowledge.

\section{Real-Time Prediction}

We apply the complete reservoir computer--the autonomous reservoir and synchronous input and output layers--to the task of predicting a chaotic time-series. To quantify the performance of our prediction algorithm, we compute the normalized root-mean-square error (NRMSE) over one Lyapunov time $T$, where $T$ is the inverse of the largest Lyapunov exponent. The $NRMSE_T$ is therefore defined as

\begin{equation}
NRMSE_T = \sqrt{\frac{\sum_{t=0}^T(u(t)-v(t))^2}{T \sigma^2}},
\end{equation}
where $\sigma^2$ is the variance of $u(t)$. 

	To train the reservoir computer, the reservoir is initially driven with the stored values of $u(t)$ as described in Sec. III and the reservoir response is recorded. This reservoir response is then transferred to a host PC. The output weights $\textbf{W}_{out}$ are chosen to minimize
    
\begin{equation}
\sum_{t=-T_{train}}^0 (u(t)-v(t))^2 + r |\textbf{W}_{out}|^2,
\end{equation}
where $r$ is the ridge regression parameter and is included in Eq. 6 to discourage over-fitting to the training set. The value of $r$ is chosen by leave-one-out cross validation on the training set. We choose a value of $T_{train}$ so that 1,500 values of $u(t)$ are used for training.

\subsection{Generation of the Mackey-Glass System}

The Mackey-Glass system is described by the time-delay differential equation

\begin{equation}
\dot{u}(t) = \beta \frac{u(t-\tau)}{1+u^n(t-\tau)} - \gamma u(t),
\end{equation}
where $\beta, \gamma, \tau,$ and $n$ are positive, real constants. The Mackey-Glass system exhibits a range of ordered and chaotic behavior. A commonly chosen set of parameters is $\beta=0.2, \gamma=0.1, \tau=17, n=10$ for which Eq. 7 exhibits chaotic behavior with an estimated largest Lyapunov exponent of 0.0086 $(T=116)$.

	Equation 10 is integrated using a $4^{th}$-order Runge-Kutta method, and the resulting series is normalized by shifting by $-1$ and passing $u(t)$ through a hyperbolic tangent function as in Ref. [11], resulting in a variance $\sigma^2=0.046$. As noted in Sec. III, $u(t)$ must be discretized according to Fig. 3b. We find an optimal temporal sampling of $\Delta t = 5$ as in Fig. 3a. 

\section{Results Analysis}
    
\begin{figure*}[!ht]
\centering
\begin{subfigure}[b]{\textwidth}
\captionsetup{font=sf}
\centering
\includegraphics[width=\textwidth]{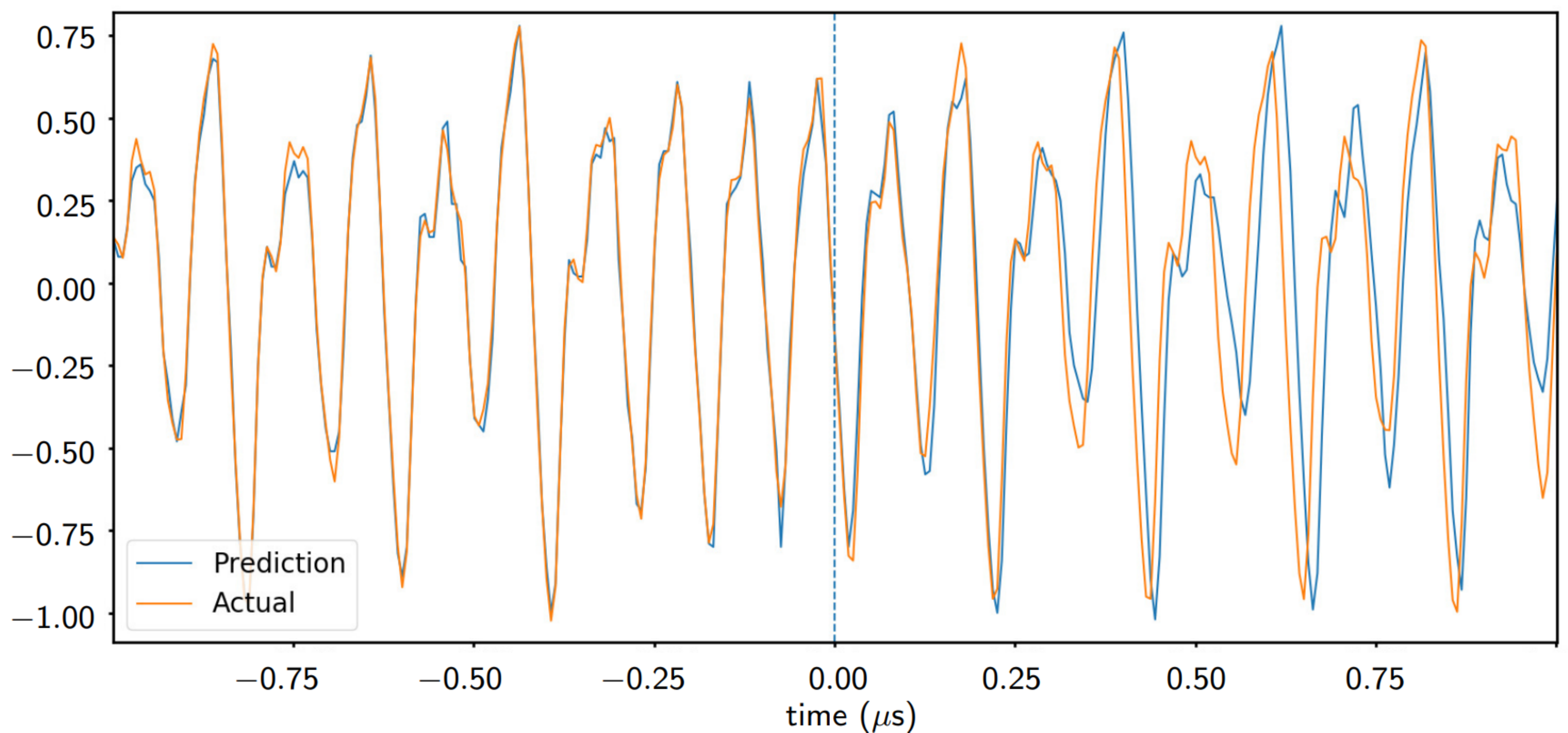}
\end{subfigure}

\caption{An example of the output of a trained reservoir computer. Autonomous generation starts at $t=0$. The target signal is the state of the Mackey-Glass system described by Eq. 11. The particular hyperparameters are $(\rho, k, \bar{\tau}, \sigma) = (1.5, 2, 11$ ns, 0.5).}
\end{figure*} 

	The reservoirs considered here are constructed from random connection matrices $\textbf{W}$ and $\textbf{W}_{in}$. However, we seek to understand the reservoir properties as functions of the hyperparameters that control the distributions of these random matrices. Recall from Sec. III that these hyperparameters are:
\begin{itemize}
\item the largest absolute eigenvalue of $\textbf{W}$, denoted by $\rho$;
\item the fixed in-degree of each node, denoted by $k$;
\item the mean delay between nodes, denoted by $\bar{\tau}$;
\item and the number of nodes which receive the input signal, denoted by $\sigma$.
\end{itemize}
Because $t_{sample}$ and, consequently, the global temporal properties of the predicting reservoir are coupled to the network size $N$, we fix $N=100$ and consider the effects of varying the four hyperparameters given above.

	Obviously, many instances of $\textbf{W}_{in}$ and $\textbf{W}$ have the same hyperparameters. We therefore consider the dynamical properties considered in this section as well as prediction performance to be random variables whose mean and variance we wish to investigate. For each set of reservoir parameters, 5 different reservoirs are created and each tested 5 times at the prediction task.
	
    For optimal choice of reservoir parameters $(\rho, k, \bar{\tau}, \sigma) = (1.5, 2, 11$ ns, 0.5), we measure $NRMSE =  0.028 \pm 0.010$ over one Lyapunov time. The predicted and actual signal trajectories for this reservoir are in Fig. 4. For comparison to other works, we prepared in ESN as in Ref. [11] with the same network size (100 nodes) and training length (1500 samples) and find a $NRMSE_T = 0.057 \pm 0.007$.

\subsection{Spectral Radius}

	The spectral radius $\rho$ controls the scale of the weights $\textbf{W}$. Though there are many ways to control this scale (such as tuning the bounds of the uniform distribution\cite{busing2010}), $\rho$ is often seen to useful way to characterize a classical ESN.\cite{caluwaerts2013spectral,lukovsevivcius2012practical} Optimizing this parameter has been critical in many applications of RC, with a spectral radius near 1 being a common starting point. More abstractly, the memory capacity has been demonstrated to be maximized at $\rho = 1.0$ from numerical experiments\cite{verstraeten2007experimental} and it has been shown that ESNs do not have the fading memory property for all inputs for $\rho > 1.0$.\cite{jaeger2001}
    
    It is not immediately clear that $\rho$ will be a similarly useful characterization of our Boolean networks, since the activation function (see Eq. 1) is discontinuous and includes time-delays--both factors which are typically not assumed to be true in the current literature. Nonetheless, we proceed with this scaling scheme and investigate the decay times and prediction performance properties of our reservoirs as we vary this parameter.
    
    We see from Fig. 5 that the performance on the Mackey-Glass prediction task is indeed optimized at $\rho=1.0$. However, performance is remarkably flat, quite unlike more traditional ESNs. The performance will obviously fail as $\rho \rightarrow 0$ (corresponding to no recurrent connections) and as $\rho \rightarrow \infty$ (corresponding to no input connections), and it appears that a range of $\rho$ in between yield similar performance. 

	This flatness in prediction performance is reflected in measures of the dynamics of the reservoir as seen in Fig. 5a and 5b. Note that the decay time of the reservoir decreases for smaller $\rho$. This is behavior is expected, because, as the network becomes more loosely self-coupled, it is effectively more strongly coupled to the input signal, and thus will more quickly forget previous inputs. More surprising is the flatness beyond $\rho=1.0$, which mirrors flatness in the performance error in this region of spectral radii. 
    
    We propose that this insensitivity to $\rho$ is due to the nature of the activation function in Eq. 3. Note that, because of the flat regions of the Heaviside step function and the fact that the Boolean state variables take discrete values, there exists a range of weights that correspond to precisely the same $\Lambda_i$ for a given node. Thus, the network dynamics are less sensitive to the exact tuning of the recurrent weights than in an ESN.
    
\begin{figure}[h]
\centering
\begin{subfigure}[b]{0.4\textwidth}
\captionsetup{font=sf}
\centering
\caption{}
\includegraphics[width=\textwidth]{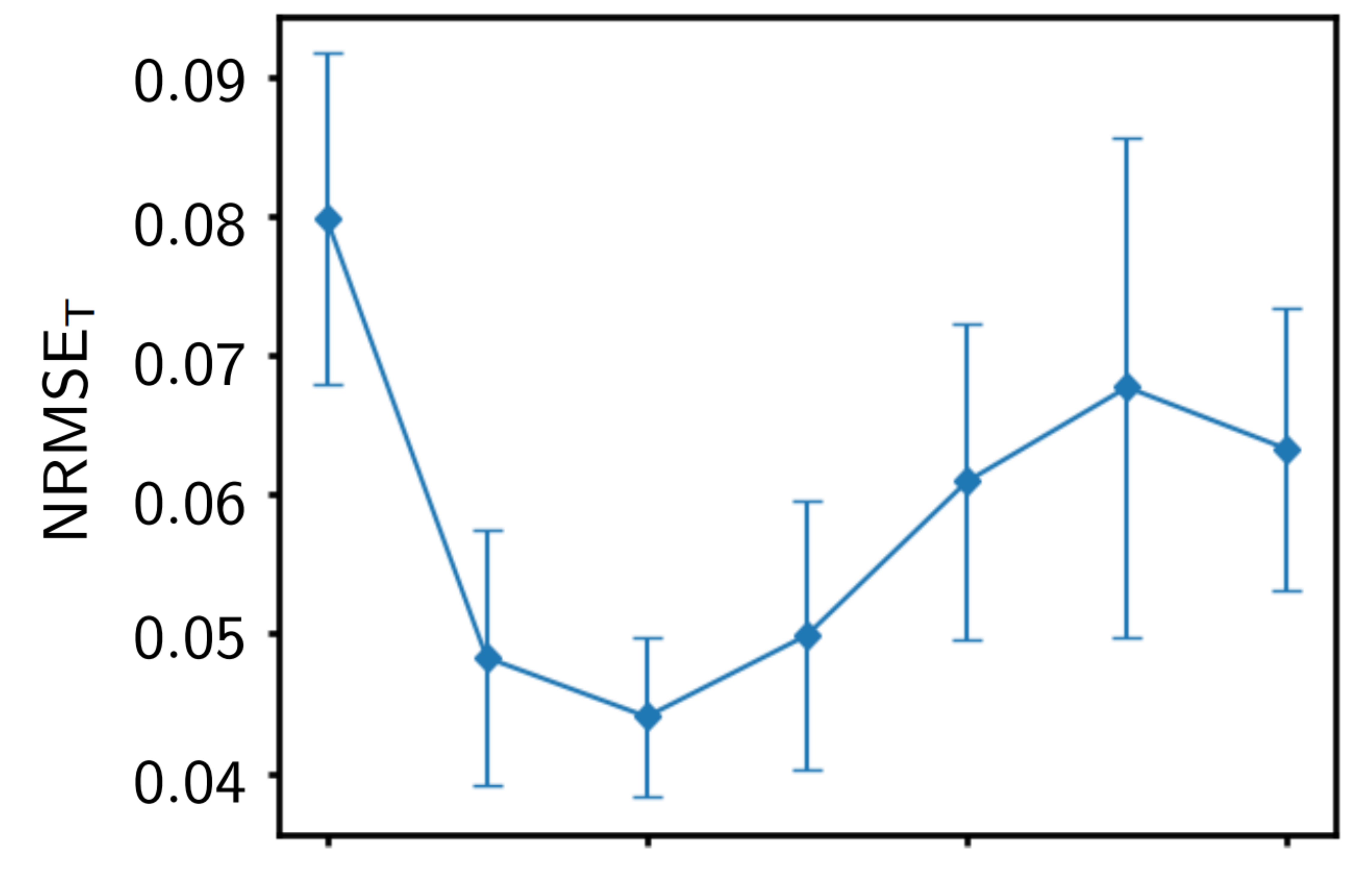}
\end{subfigure}
\centering
\begin{subfigure}[b]{0.4\textwidth}
\captionsetup{font=sf}
\centering
\caption{}
\includegraphics[width=\textwidth]{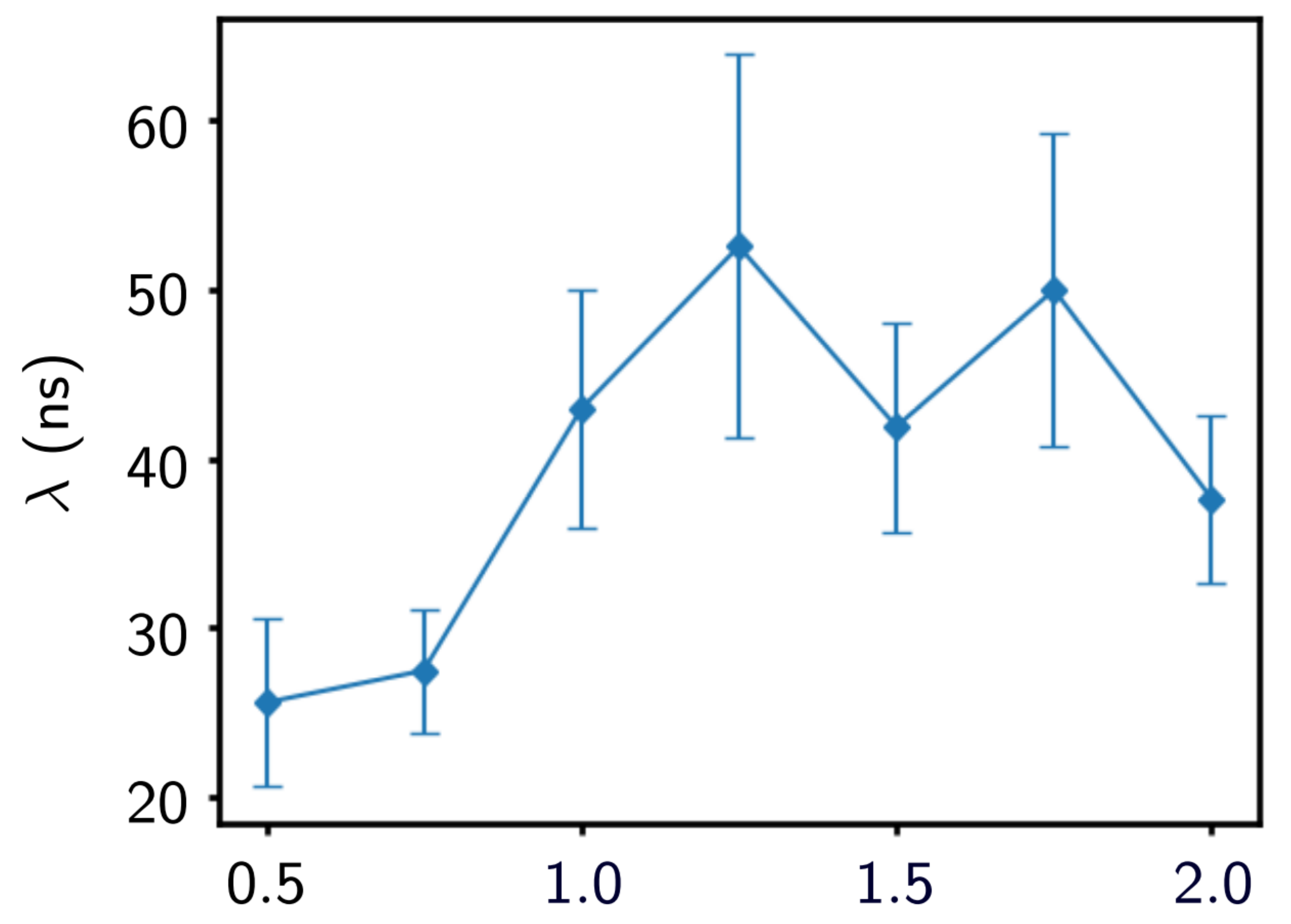}
\end{subfigure}
\caption{Prediction performance and fading memory of reservoirs with $(k, \bar{\tau}, \sigma) = ($2, 11 ns, 0.75) and varying $\rho$. (a) Somewhat consistent with observations in echo-state networks, $\rho$ near 1.0 appears to be a good choice. However, a much wider range of $\rho$ suffice as well. (b) As $\rho$ becomes small and the reservoir becomes more strongly coupled to the input, the reservoir more quickly forgets previous inputs. The decay time levels out above $\rho=1.0$. Note that $\lambda$ is everywhere the same order of magnitude as $\bar{\tau}$.}
\end{figure}    

\subsection{Connectivity}
    
	The second component to characterizing $\textbf{W}$ is the in-degree $k$ of the nodes, which is the density of non-zero entries in the row vectors of $\textbf{W}$. Because the $\Lambda_i$'s are populated by explicit calculation of the functions in Eq. 3 and because larger $\Lambda_i$'s require more resources to realize in hardware, it is advantageous to limit $k$. We therefore ensure that each node has fixed $k$ rather than simply some mean degree that is allowed to vary.
    
    From the study of purely Boolean networks with discrete-time dynamics (\textit{i.e.}, dynamics defined by a map rather than a differential equation), a transition from order to chaos is seen in a number of network motifs at $k=2$.\cite{derrida1986random,rohlf2002criticality} In fact, Hopefield type nodes are seen to have this critical connectivity in the explicit context of RC.\cite{busing2010} The connectivity is a commonly optimized hyperparameter in the context of ESNs as well\cite{jaeger2001, jaeger2002tutorial} with the common heuristic that low-connectivity $(1-5\%$ of $N)$ promotes a richer reservoir response. 
    
    From the above considerations, we study the reservoir dynamics and prediction performance as we vary $k=1-4.$ From Fig. 6, we see stark contrasts from the picture of RC with a Boolean network in discrete time. First, the reservoirs remain in the ordered phase for $k=2-4$, which clearly demonstrates that the real-valued nature of the underlying dynamical variables in Eq. 3 are critically important to the network dynamics.
    
    We see further in Fig. 6b that the mean decay time increases with increasing $k$, \textit{i.e.}, that the network takes longer to forget past inputs when the nodes are more densely connected. This phenomenon is perhaps understood by the increased number of paths in networks with higher $k$. These paths provide more avenues for information about previous network states to propagate, thus prolonging the decay of the difference in Eq. 6. The variance in decay time also significantly increases for increasing $k$. This may be an indicator of eventual criticality for large enough $k$.
    
    Given the strong differences in reservoir dynamics between $k=1,4$, it is surprising that no significant difference at the prediction task is detected. However, it is useful for the design of efficient reservoirs to observe that very sparsely connected reservoirs suffice for complicated tasks. As noted in Sec. IV, nodes with more inputs require more resources to realize in hardware and more processing time to compute the corresponding $\Lambda_i$ in Eq. 3. 
    
\begin{figure}[h]
\centering
\begin{subfigure}[b]{0.4\textwidth}
\captionsetup{font=sf}
\centering
\caption{}
\includegraphics[width=\textwidth]{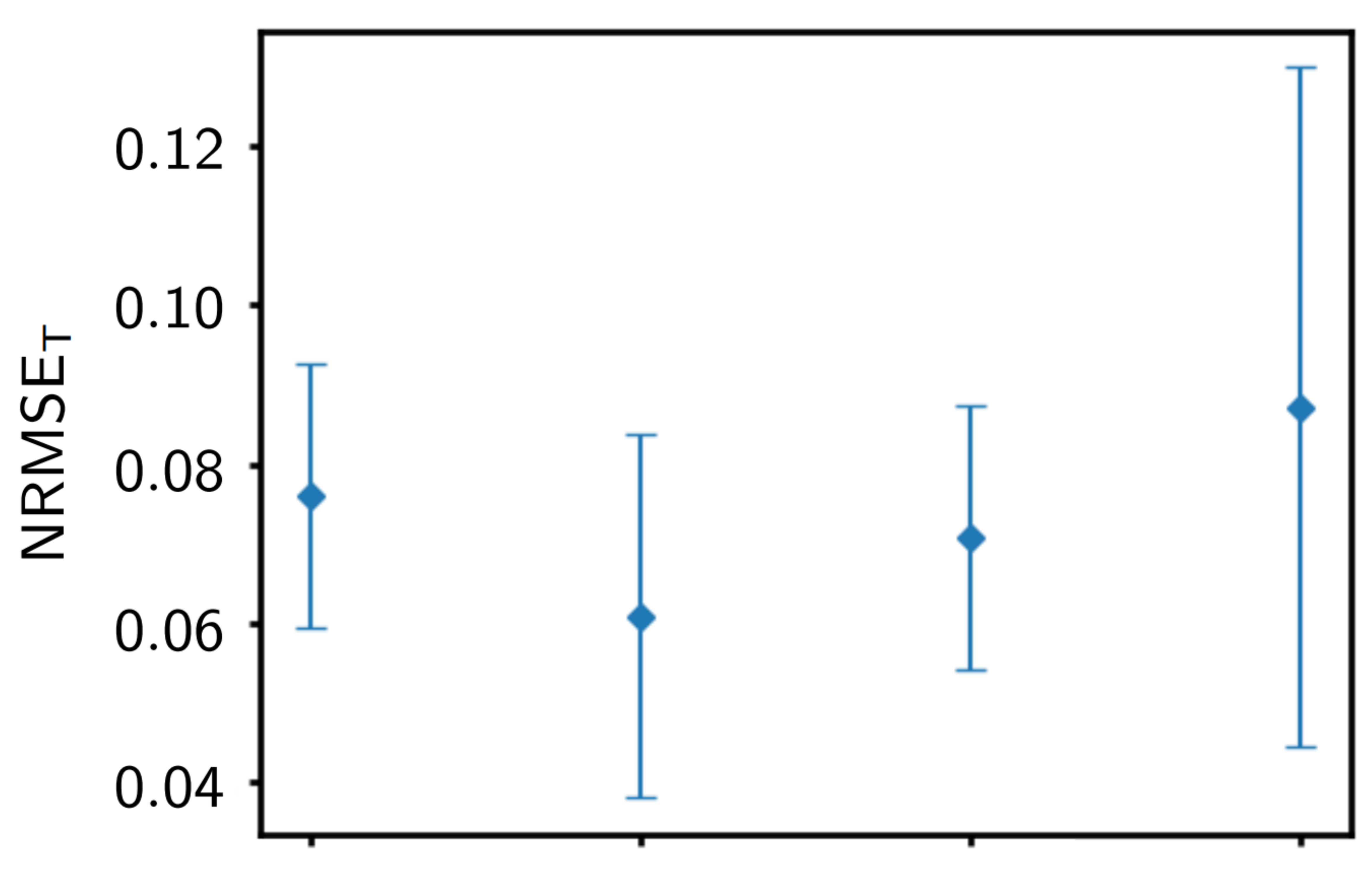}
\end{subfigure}
\centering
\begin{subfigure}[b]{0.4\textwidth}
\captionsetup{font=sf}
\centering
\caption{}
\includegraphics[width=\textwidth]{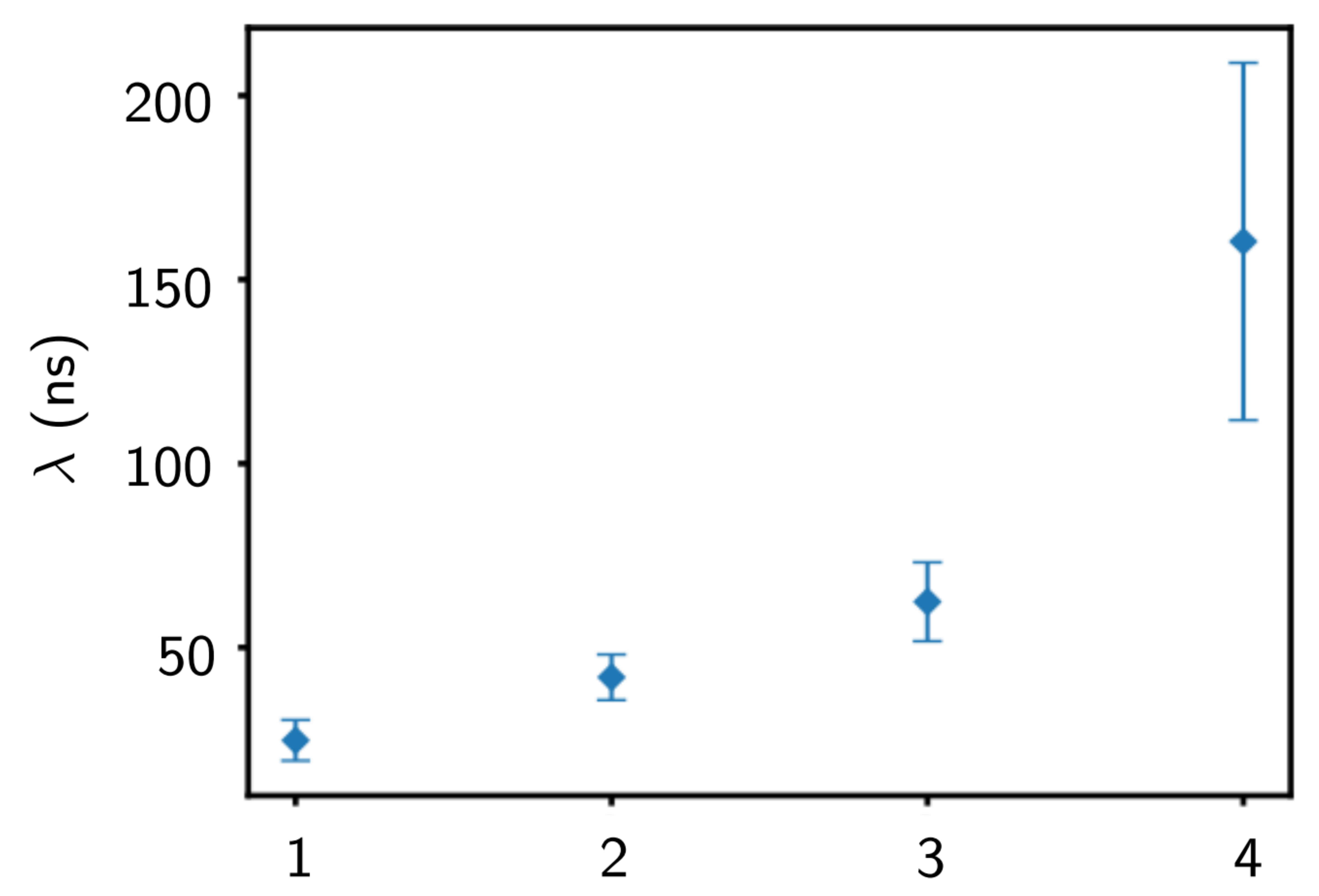}
\end{subfigure}
\caption{Prediction performance and fading memory of reservoirs with $(\rho, \bar{\tau}, \sigma) = ($1.5, 11 ns, 0.75) and varying $k$. (a) We see effectively no difference over this range, contrary to intuitions from studies of Boolean networks in discrete time. (b) For $k=1$, $\lambda$ is approximately equal to $\bar{\tau}$. However, as we increase $k$ to 4, both the mean and variance of $\lambda$ approaches almost an order of magnitude larger than $\bar{\tau}$.}
\end{figure} 
    
\subsection{Mean Delay}

	As argued in Sec. III, adding time-delays along the network links increases the characteristic time scale of the network. We distribute delays by randomly choosing, for each network link, a delay time from a uniform distribution from $\bar{\tau}/2-3\bar{\tau}/2$. The shape of this distribution is chosen to fix the mean delay time while keeping the minimum delay time above the characteristic time of the nodes themselves.
    
    In Fig. 7 we compare the prediction performance vs. $\bar{\tau}$. Note that this parameter is most critical in achieving good prediction performance in the sense that $\bar{\tau}$ being comparable to $\tau_{node}$ yields poor performance. However, the performance is flat past a certain minimum $\bar{\tau}$ near 8.5 ns. This point is important to identify, as adding more delay elements than necessary increases the number of FPGA resources needed to realize the network. 

\begin{figure}[h]
\centering
\begin{subfigure}[b]{0.4\textwidth}
\captionsetup{font=sf}
\centering
\includegraphics[width=\textwidth]{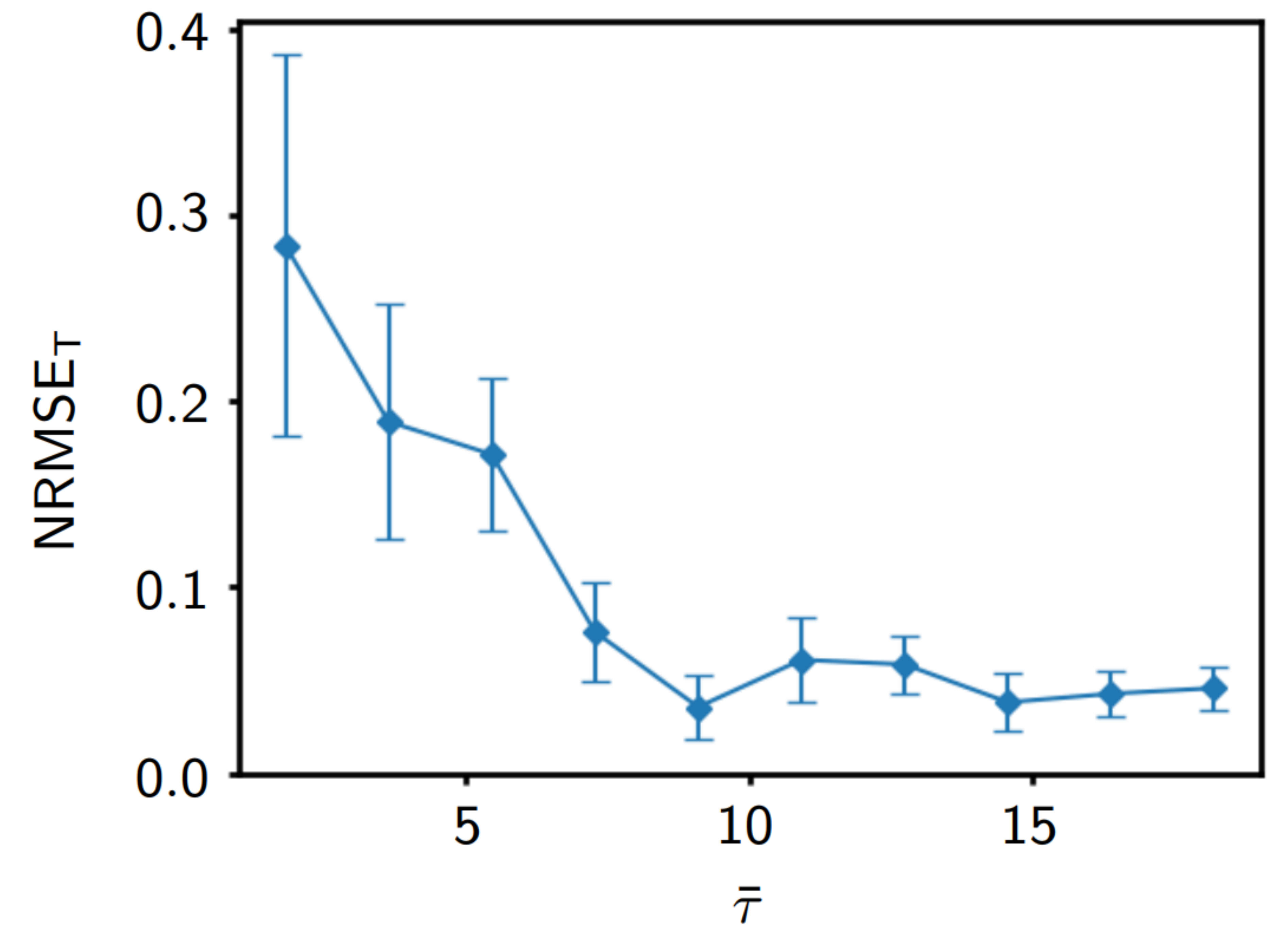}
\end{subfigure}
\caption{Prediction performance of reservoirs with $(\rho, k, \sigma) = ($1.5, 2, 0.75) and varying $\bar{\tau}$. The NRMSE decreases until approximately $\bar{\tau}=9.5$, after which point it remains approximately constant.}
\end{figure} 
    
\subsection{Input Density}
    
    We finally consider the effect of tuning the proportion of reservoir nodes that are connected to the input signal. This proportion is often assumed to be 1,\cite{jaeger2002tutorial} although recent studies have shown a smaller fraction to be useful in certain situations, such as predicting the Lorenz system.\cite{pathak2018hybrid}
    
    We observe from Fig. 8a that an input density of 0.5 performs better than input densities of 0.25, 0.75, and 1.0. We note from Fig. 8b that this corresponds to the point of longest decay time. The decreasing decay time with higher input densities 0.75 and 1.0 are consistent with the expectation that reservoirs that are more highly coupled to the input signal will forget previous inputs more quickly. 
    
    It is apparent from Fig. 8b that the input density is a useful characterization of the RC scheme, impacting the fading memory properties of the reservoir-input system and ultimately improving performance by a factor of 3 when compared to a fully dense input matrix. This results suggests the input density to be a hyperparameter deserving of more attention in general contexts.
    
\begin{figure}[h]
\centering
\begin{subfigure}[b]{0.4\textwidth}
\captionsetup{font=sf}
\centering
\caption{}
\includegraphics[width=\textwidth]{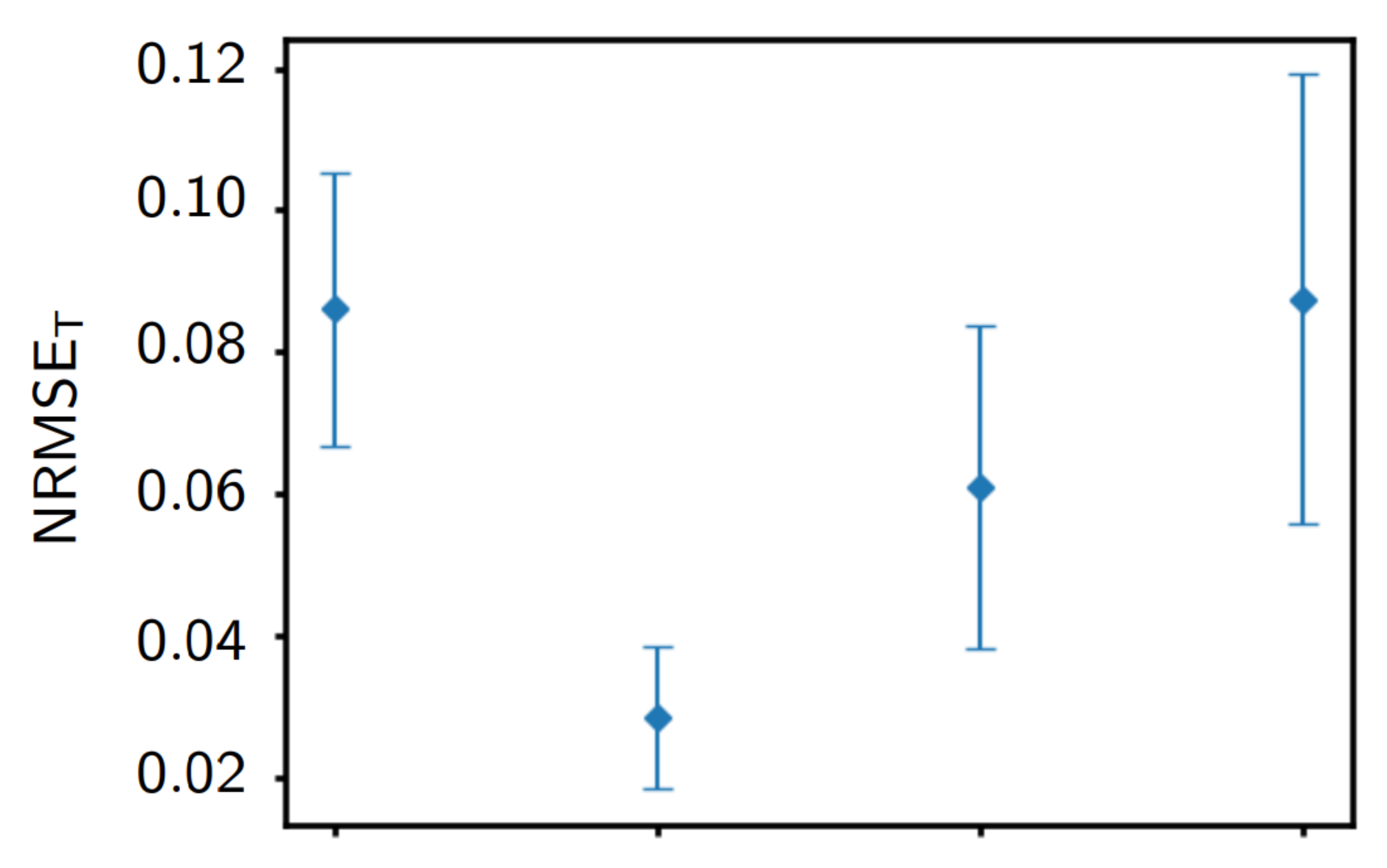}
\end{subfigure}
\centering
\begin{subfigure}[b]{0.4\textwidth}
\captionsetup{font=sf}
\centering
\caption{}
\includegraphics[width=\textwidth]{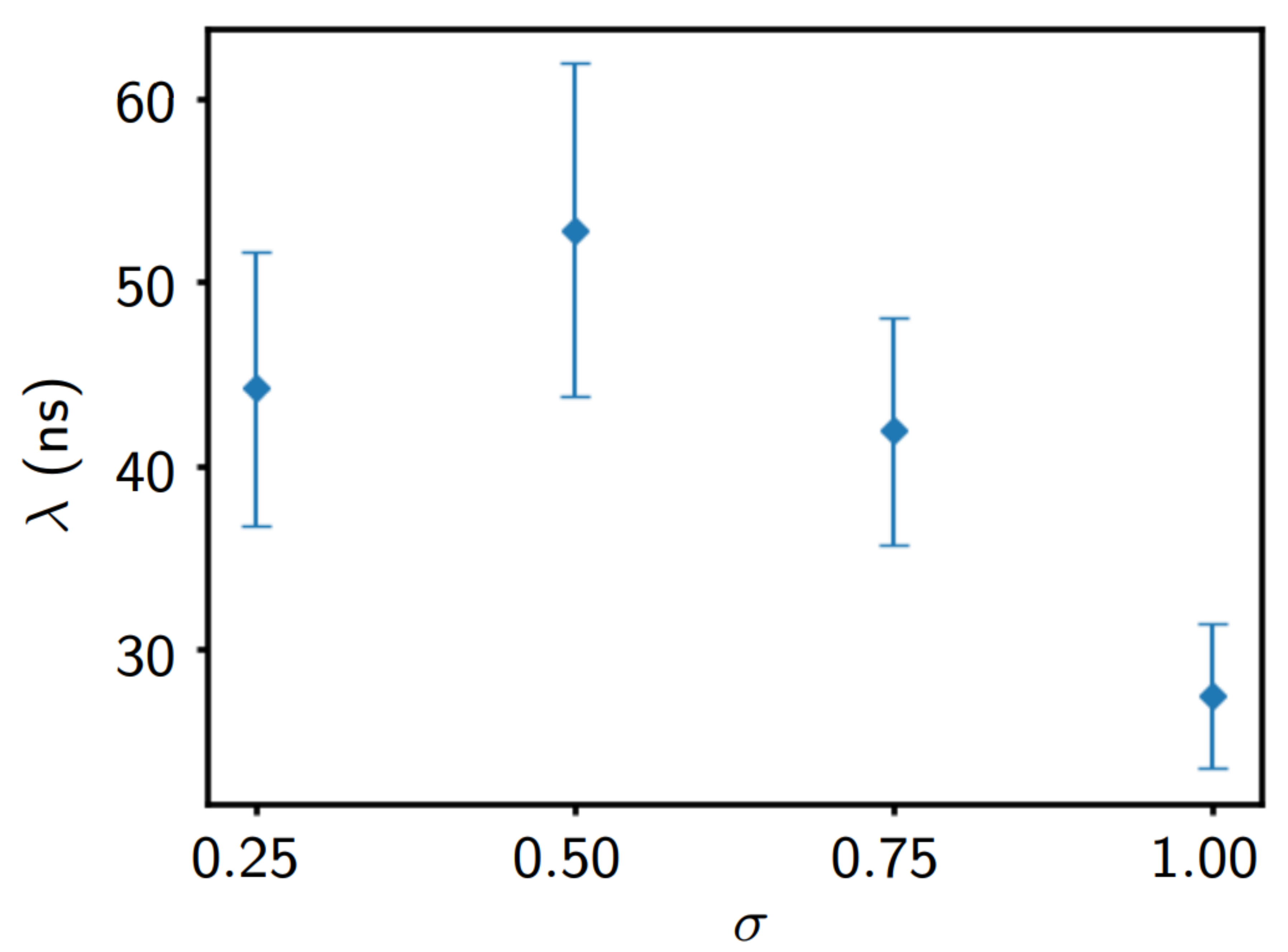}
\end{subfigure}
\caption{Prediction performance and fading memory of reservoirs with $(\rho, k, \bar{\tau}) = ($1.5, 2, 11 ns, 0.75) and varying $\rho$. (a) Choosing $\sigma = 0.5$ improves prediction performance by a factor of 3 over the usual choice of $\sigma = 1.0$ (b) With larger $\sigma$, the reservoir is more strongly coupled to the input signal. Consequently, $\lambda$ decreases, signifying that the reservoir is more quickly forgetting previous inputs.}
\end{figure} 
    
\subsection{Attractor Reconstruction}
    
    Prediction algorithms are commonly evaluated on their short-term prediction abilities as we have done so far in this section. The predicted and actual signal trajectories will always diverge in the presence of chaos due to the positivity of at least one Lyapunov exponent. However, it has been seen recently that reservoir computers\cite{pathak2017} and other neural network prediction schemes\cite{qiao2017deep} can have similar long-term behavior as the target system. In particular for ESNs, it has been seen that different reservoirs can have similar short-term prediction capabilities, but very different long-term behavior, with some reservoirs capturing the climate of the Lorenz system and others eventually collapsing onto a non-chaotic attractor.\cite{pathak2017} This phenomenon has recently been explained in terms of generalized synchronization--a stronger condition than fading memory.\cite{lu2018}
    
    To observe a similar phenomenon in the RC scheme considered here, we allow a trained reservoir to evolve for 100 Lyapunov times (about $15$ $\mu$s) beyond the training period. The last half of this period is visualized in time-delay phase-space to see if the climate of the true Mackey-Glass system is replicated. 
    
    Our results show phenomena consistent with previous observations in ESNs. Figure 9a shows the true attractor of Eq. 11, which has fractal dimension and is non-periodic. Figure 9b shows the attractor of a well-chosen autonomous, Boolean reservoir. Although the attractor is ``fuzzy," the trajectory remains on a Mackey-Glass-like shape well beyond the training period. On the other hand, a reservoir with similar short-term prediction error is shown in Fig. 9c. Although this network is able to replicate the short-term dynamics of Eq. 11, its attractor is very unlike the true attractor in Fig. 9a. This results shows that, even in the presence of noise inherent in physical systems, the autonomous Boolean reservoir can learn the long-term behaviors of a complicated, chaotic system.
    
\begin{figure}[h]
\centering
\begin{subfigure}[b]{0.23\textwidth}
\captionsetup{font=sf}
\centering
\caption{}
\includegraphics[width=\textwidth]{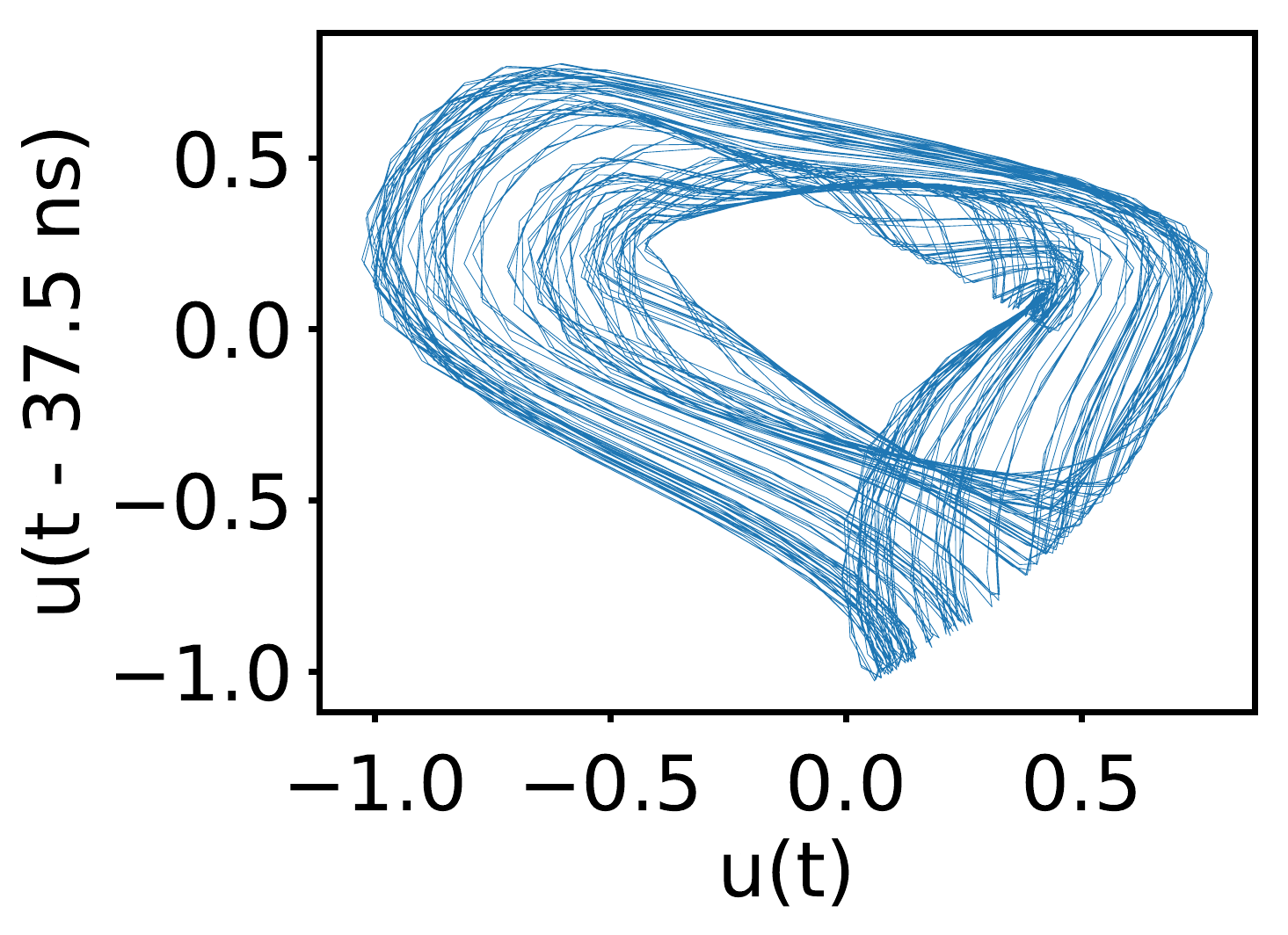}
\vspace{-9mm}
\end{subfigure}
\centering
\begin{subfigure}[b]{0.23\textwidth}
\captionsetup{font=sf}
\centering
\caption{}
\includegraphics[width=\textwidth]{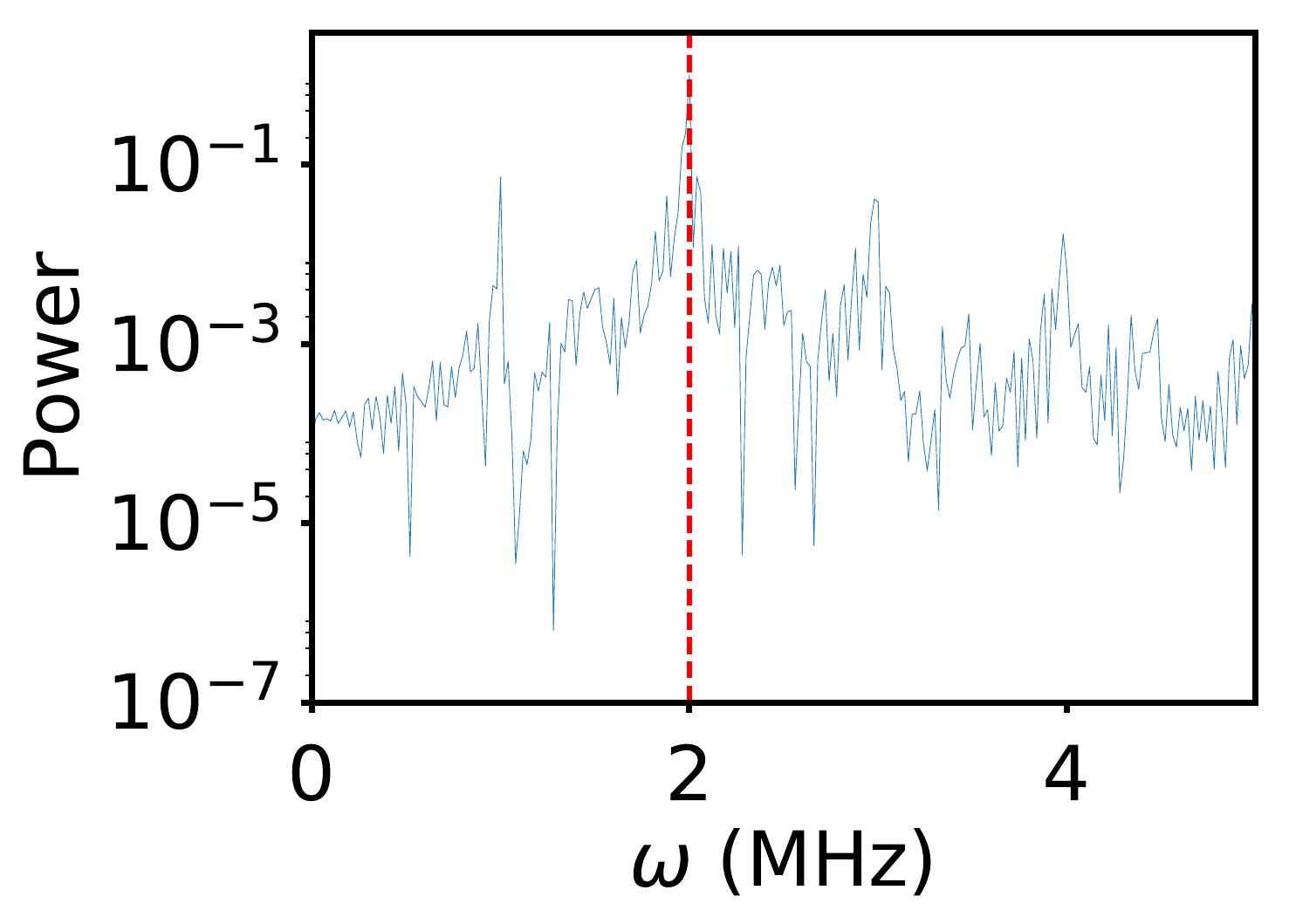}
\vspace{-9mm}
\end{subfigure}
\centering
\begin{subfigure}[b]{0.23\textwidth}
\captionsetup{font=sf}
\centering
\caption{}
\includegraphics[width=\textwidth]{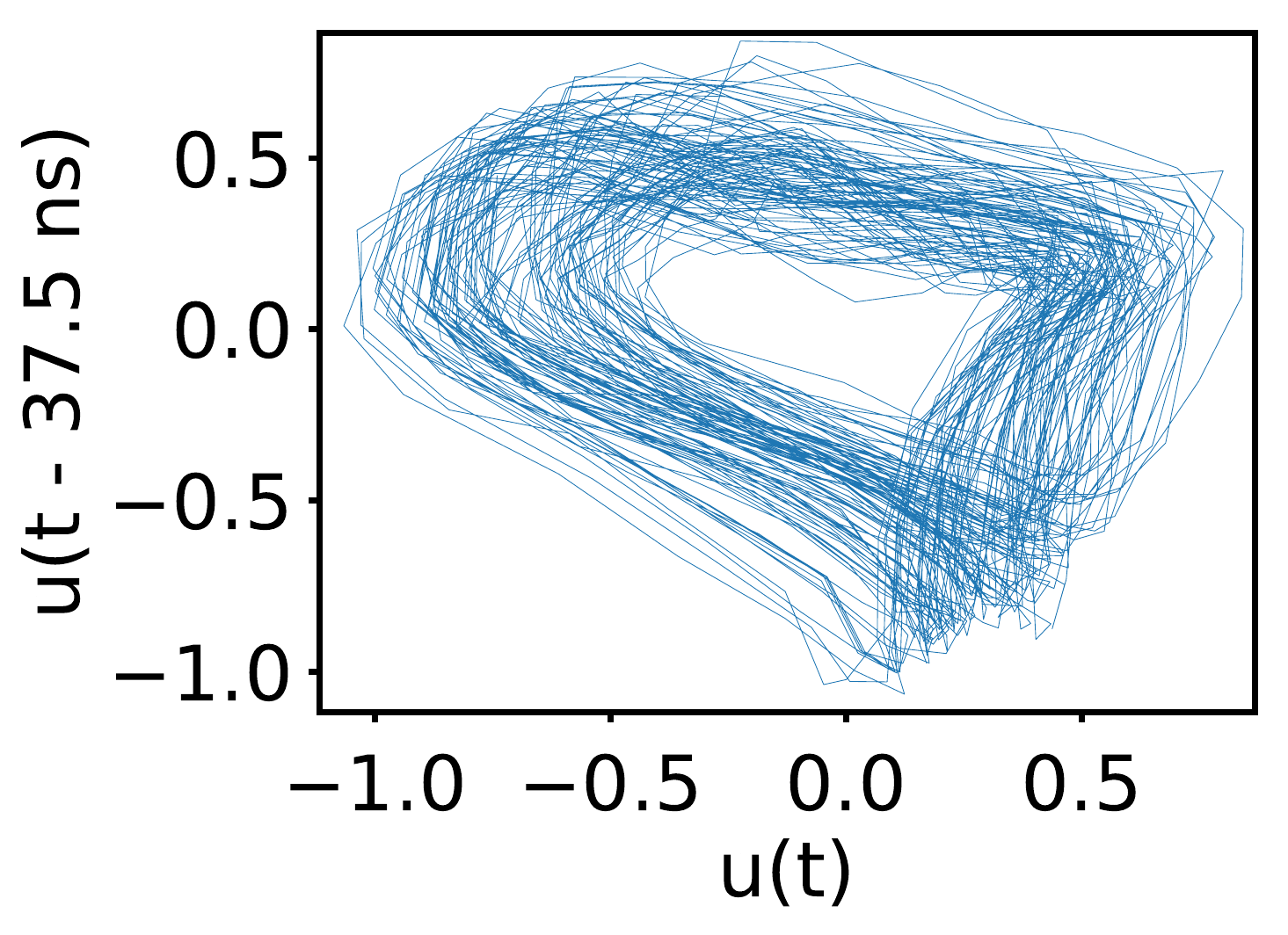}
\vspace{-9mm}
\end{subfigure}
\centering
\begin{subfigure}[b]{0.23\textwidth}
\captionsetup{font=sf}
\centering
\caption{}
\includegraphics[width=\textwidth]{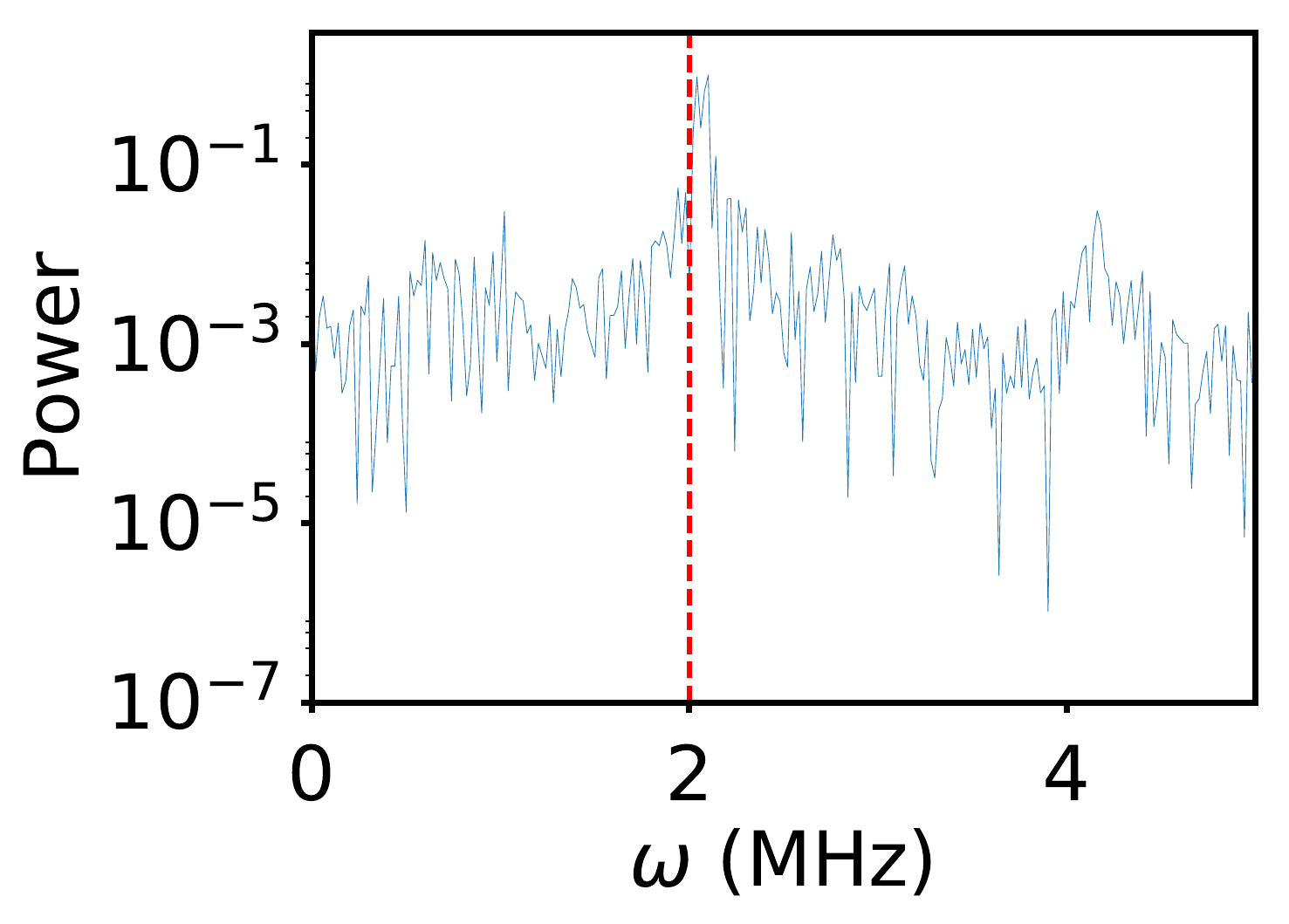}
\vspace{-9mm}
\end{subfigure}
\centering
\begin{subfigure}[b]{0.23\textwidth}
\captionsetup{font=sf}
\centering
\caption{}
\includegraphics[width=\textwidth]{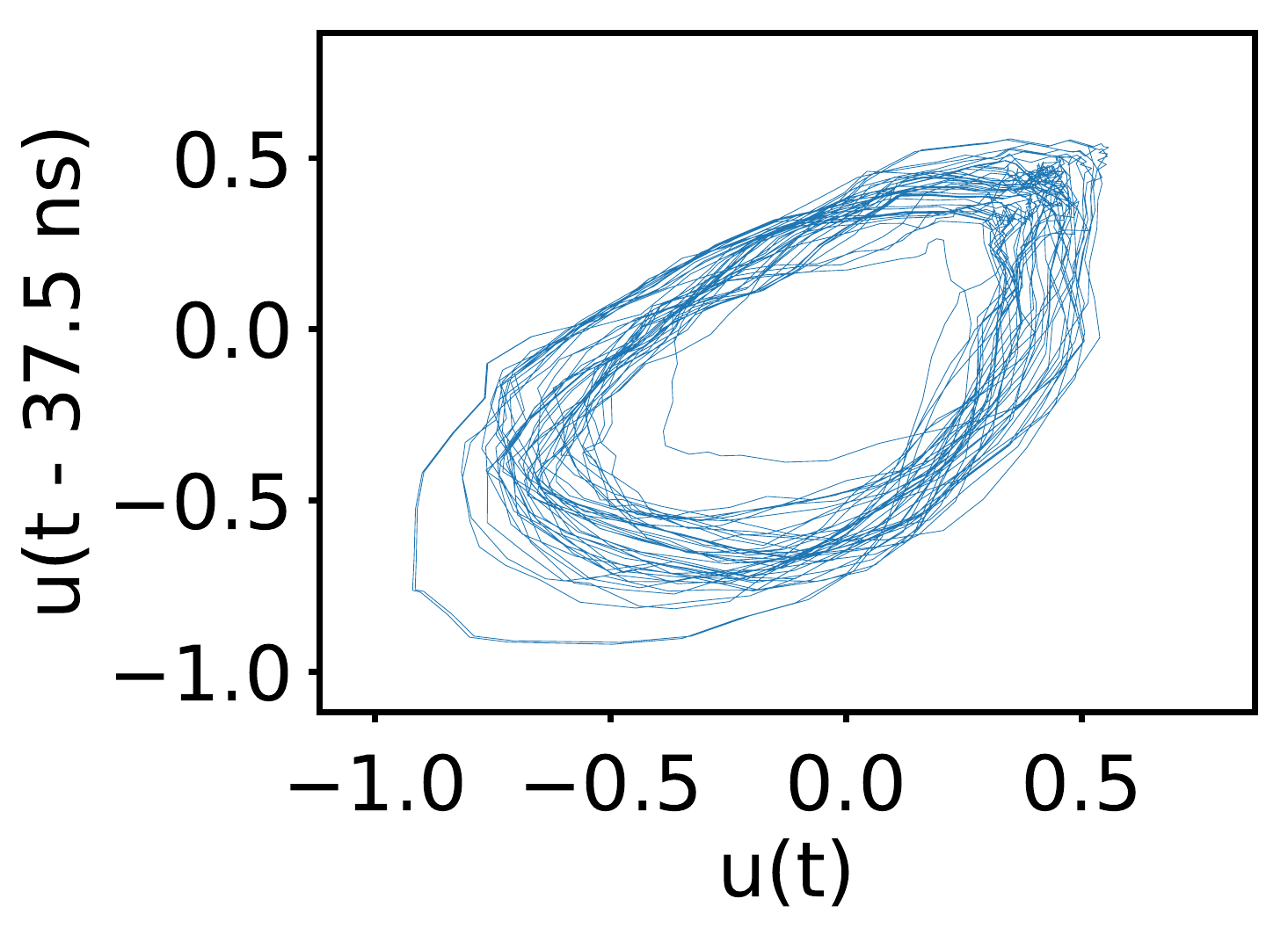}
\end{subfigure}
\centering
\begin{subfigure}[b]{0.23\textwidth}
\captionsetup{font=sf}
\centering
\caption{}
\includegraphics[width=\textwidth]{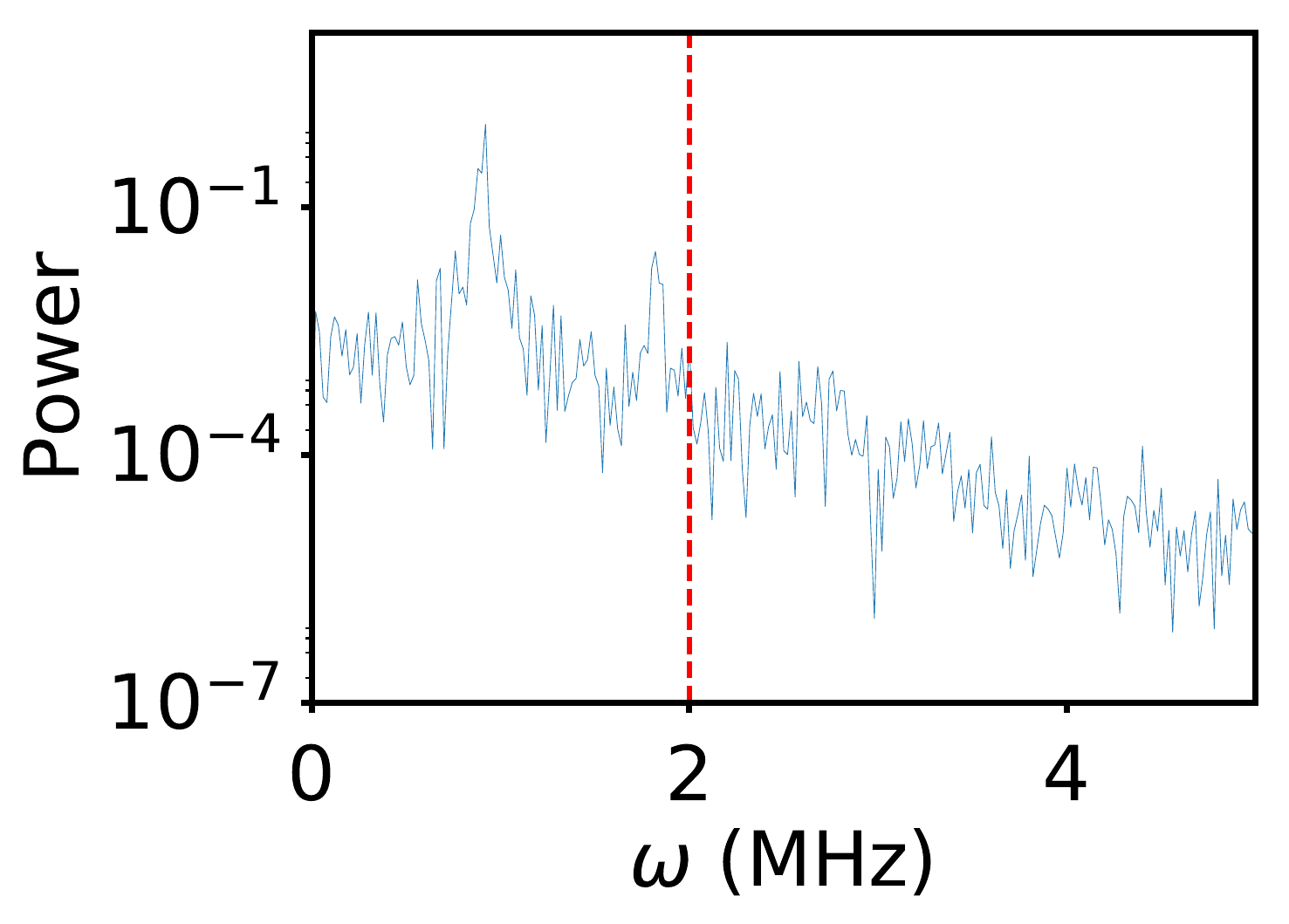}
\end{subfigure}
\caption{Phase-space representations and power spectra of the attractors of Eq. 11 and trained reservoirs. (a) The true attractor and (b) normalized power spectrum of the Mackey-Glass system, as presented to the reservoir. (c) The attractor and (d) normalized power spectrum for a reservoir whose long-term behavior is similar to the true Makcey-Glass system. Although ``fuzzy," the attractor remains near the true attractor. The power spectrum shows a peak 0.10 MHz away from the true peak. The hyperparameters for this reservoir are $(\rho, k, \bar{\tau}, \sigma) = (1.5, 2, 11$ ns, $0.75)$. (e) The attractor and (f) normalized power spectrum of a reservoir whose long-term behavior is different than the true Mackey-Glass system. The dominate frequency of the true system is highly suppressed, while a lower-frequency mode is amplified. The hyperparameters for this reservoir are $(\rho, k, \bar{\tau}, \sigma) = (1.5, 4, 11$ ns, $0.75)$. The dashed, red line in the power spectrum plots indicates the peak of the spectrum in the true Mackey-Glass system.}
\end{figure} 

\section{Discussion and Conclusion}

	We conclude that an autonomous, time-delay, Boolean network serves as a suitable reservoir for RC. We have demonstrated that such a network can perform the complicated task of predicting the evolution of a chaotic dynamical system with comparable accuracy to software-based RC. We have demonstrated the state-of-the-art speed with which our reservoir computer can perform this calculation, exceeding previous hardware-based solutions to the prediction problem. We have demonstrated that, even after the trained reservoir computer deviates from the target trajectory, the attractor stays close to the true attractor of the target system.
    
    This work demonstrates that fast, real-time computation with autonomous dynamical systems is possible with readily-available electronic devices. This technique may find applications in design of systems that require estimation of the future state of a system that evolves on a nanosecond to microsecond time scale, such as the evolution of cracks through crystalline structures or the motion of molecular proteins.

\section{Acknowledgements}
We gratefully acknowledge discussions of this work with Roger Brockett, Michele Girvan, Brian Hunt, and Edward Ott, and the financial support of U.S. Army Research Office Grant No. W911NF-12-1-0099.

\appendix

\section{Realizing the Reservoir on an FPGA}

	In this appendix, we present the hardware description code for the reservoir nodes, delay lines, and a small reservoir. The code is written in \textit{Verilog} and compiled using Altera's \textit{Quartus Prime} software. Some parts of the code depend on the number of reservoir nodes $N$, the node in-degree $k$, and the number of bits $n$ used to represent the input signal $u(t)$. We give explicitly the code only for $N=3$, $k=2$, and $n=1$, but generalizations are straightforward. 
    
    As discussed in Sec. III, reservoir nodes implement a Boolean function $\Lambda_i : \textbf{Z}_2^{k+n} \rightarrow \textbf{Z}_2$ of the form given in Eq. 3. Each Boolean function can be defined by a Boolean string of length $2^{k+n}$ that specifies the look-up-table (LUT) corresponding of the Boolean function. For example, the AND function maps $\textbf{Z}_2^2 \rightarrow \textbf{Z}_2$ and has the LUT defined in Fig. 10. The Boolean string that defines the AND function is $0001$ as can be seen from the the right-most column of the LUT.  

\begin{figure}[h]
\centering
\begin{subfigure}[b]{0.2\textwidth}
\captionsetup{font=sf}
\centering
\includegraphics[width=\textwidth]{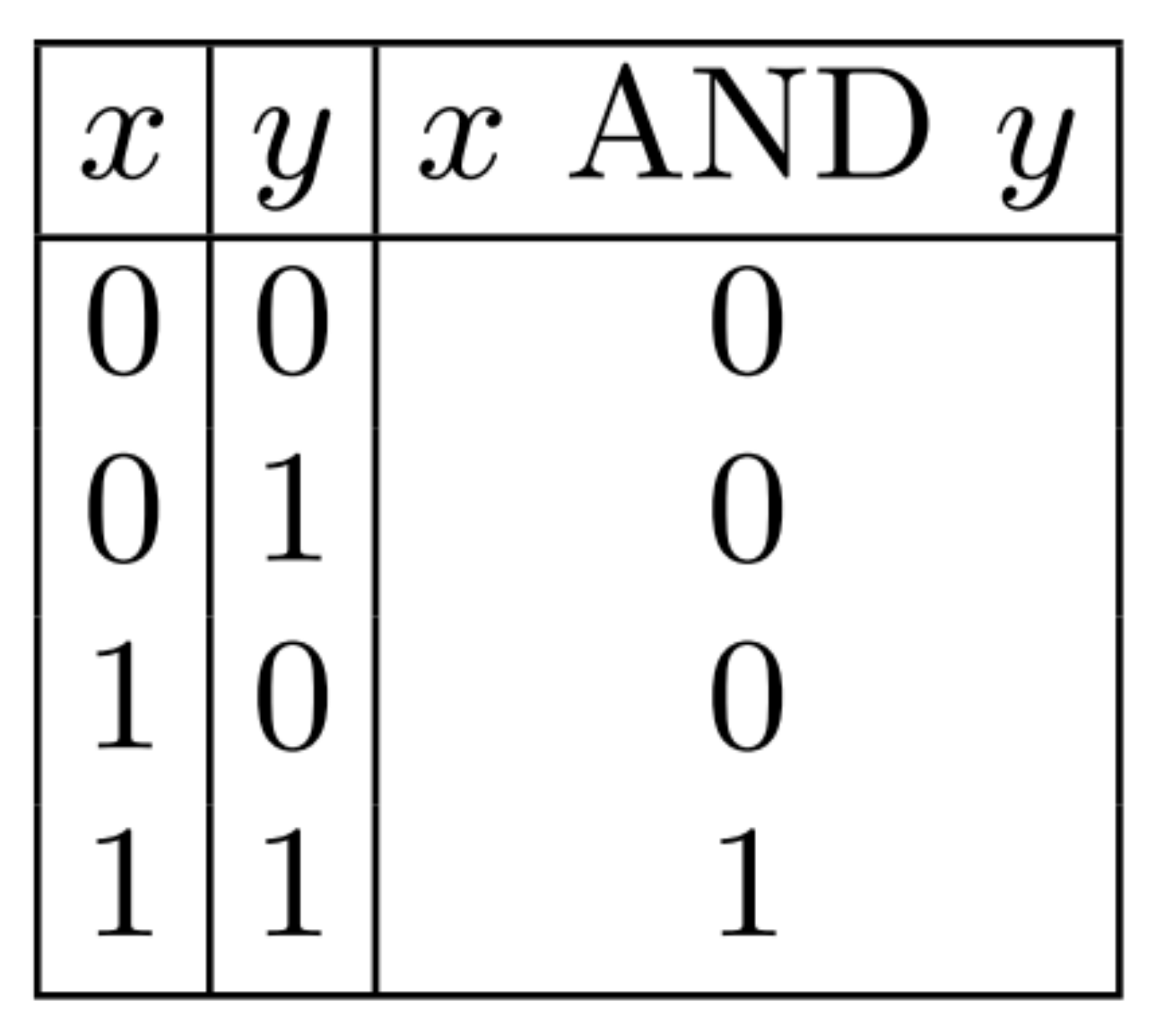}
\end{subfigure}
\caption{The LUT for the AND function. It can be specified by the Boolean string that makes up the right-most column.}
\end{figure}
    
    The code given in Fig. 11 generates a node with Boolean function based on any LUT of length $2^3=8$. The module \textbf{node} is declared in line 1 with inputs \textbf{node\_in} and output \textbf{node\_out}. The width of \textbf{node\_in} is 3 bits as specified in line 3. The parameter \textbf{lut} is declared in line 2. Note that it is initialized to some value as required by \textit{Quartus}, but this value is changed whenever a node is declared within the larger code that defines the complete reservoir.  

\begin{figure}[h]
\centering
\begin{subfigure}[b]{0.45\textwidth}
\captionsetup{font=sf}
\centering
\includegraphics[width=\textwidth]{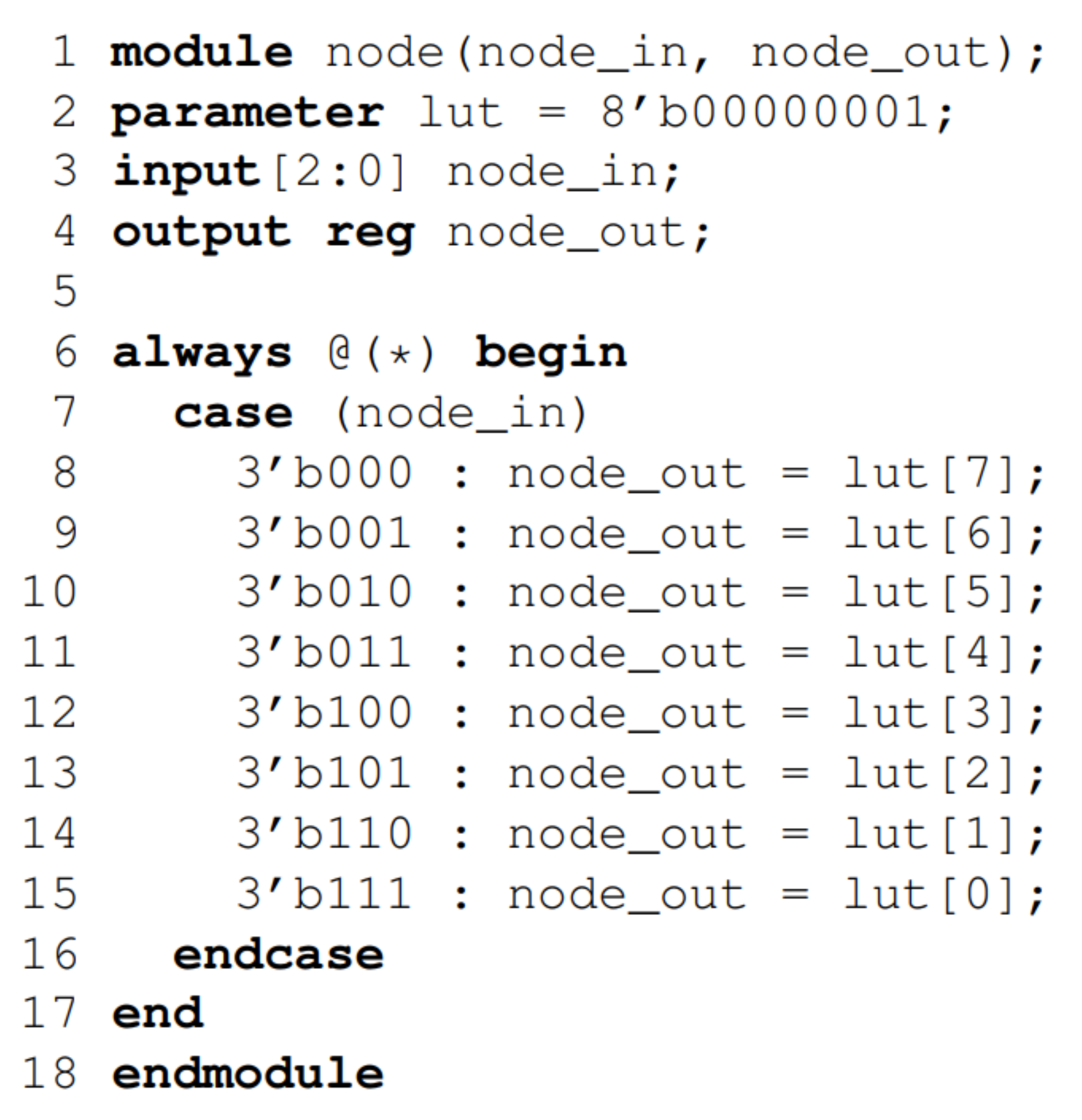}
\end{subfigure}
\caption{Verilog code for a generic node that can implement any 3-input Boolean function, specified by a Boolean string of length 8.}
\end{figure}
    
    The main part of the code is within an \textbf{always @(*)} block, which creates an inferred sensitivity list and is used to create arbitrary combinational logic. Line 7 specifies that values before the colon in the proceeding lines correspond to \textbf{node\_in}. The statement following the colon determines which value is assigned to \textbf{node\_out}. In effect, line 8 simply specifies that, whenever the value of \textbf{node\_in} is a 3-bit string equal to 000, the value of \textbf{node\_out} is whatever the value of \textbf{lut[7]} is. For example, if we create an instance of the module \textbf{node} with parameter \textbf{lut=8'b00000001}, then the node will execute the 3 input AND function. 
    
    As discussed in Sec. IV, delay lines are created as chains of pairs of inverter gates. Such a chain of length $2m$ is created with the code in Fig. 12. Similarly to the \textbf{node} module, the \textbf{delay\_line} module is declared in line 1 with the input \textbf{delay\_in} and output \textbf{delay\_out}. It has a parameter $m$ which specifies the number of pairs in the chain and can be changed when calling a specific instance of \textbf{delay\_line}. A number of wires are declared in line 5 and will be used as the inverter gates. Note the important directive \textbf{/*synthesis keep*/}, which instructs the compiler to not simplify the module by eliminating the inverter gates. This is necessary, because otherwise the compiler would realize that \textbf{delay\_line}'s function is trivial and remove all of the inverter gates.

\begin{figure}[h]
\centering
\begin{subfigure}[b]{0.45\textwidth}
\captionsetup{font=sf}
\centering
\includegraphics[width=\textwidth]{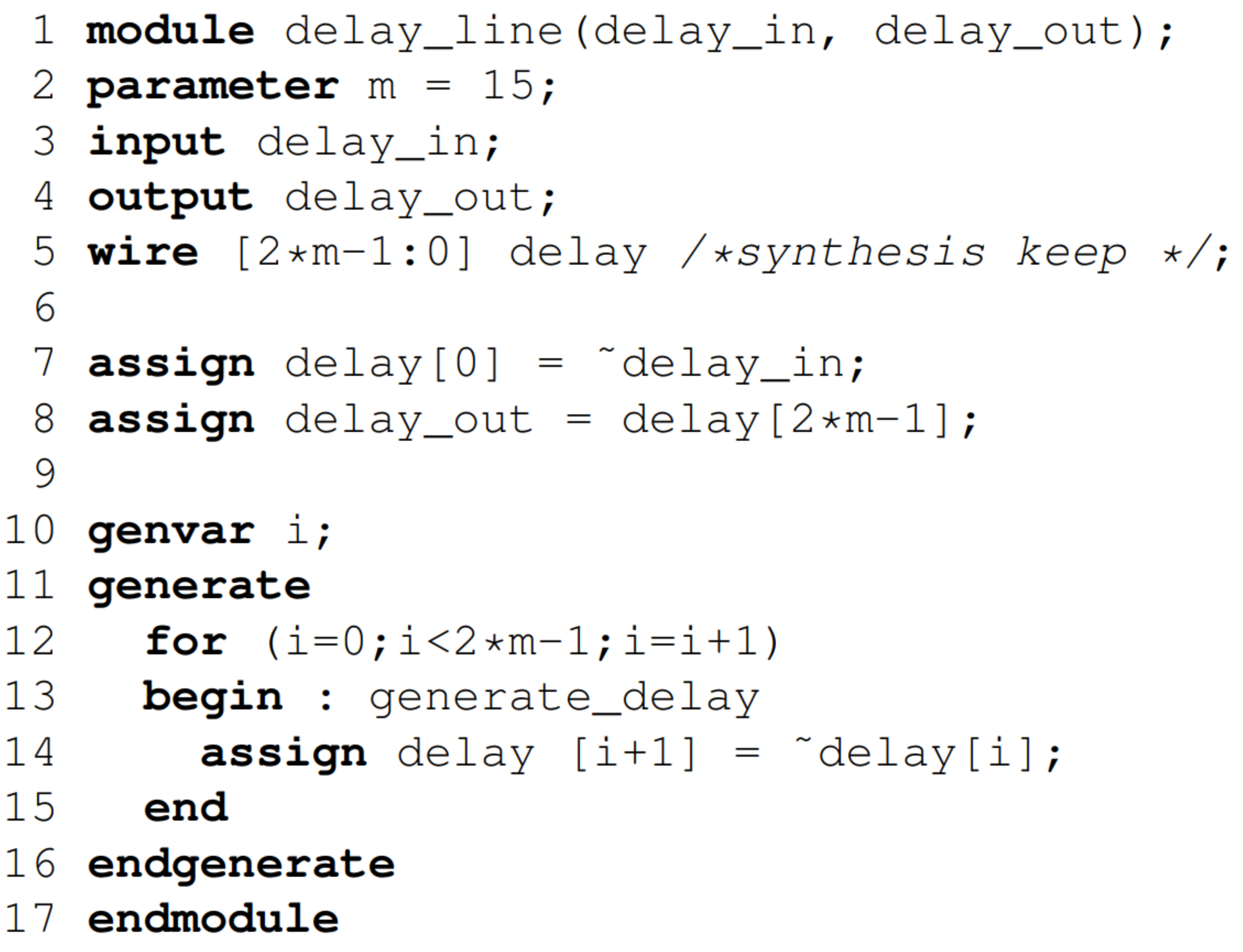}
\end{subfigure}
\caption{Verilog code for a delay line with $2m$ inverter gates.}
\end{figure}

	Lines 7-8 specify the beginning and end of the delay chain as the \textbf{delay\_in} and \textbf{delay\_out}, respectively. Lines 10-16 use a \textbf{generate} block to create a loop that places inverter gates in between \textbf{delay\_in} and \textbf{delay\_out}, resulting in a delay chain of length $2m$. 

	The \textbf{reservoir} module is the code that creates $N$ instances of \textbf{node} and connects them $Nk$ instances of \textbf{delay\_line}. As an illustrative example, consider a 3-node reservoir with the following parameters
    
\begin{equation}
\textbf{W} = 
\begin{bmatrix}
    0.1 & 0.3 & 0 \\
    -0.2 & 0 & 0.1 \\
    -0.3 & 0.2 & 0
\end{bmatrix}
\end{equation}

\begin{equation}
\textbf{W}_{in} = 
\begin{bmatrix}
    0.1 \\
    -0.2 \\
    0.2
\end{bmatrix}
\end{equation}

\begin{equation}
\bm{\tau} = 
\begin{bmatrix}
    10 & 15 & 0 \\
    6 & 0 & 7 \\
    12 & 10 & 0
\end{bmatrix}
\end{equation}
and only a 1-bit representation of $u(t)$. When we pass $u(t)$ and $x(t)$ into the \textbf{node} module, we index such that $u(t)$ comes first, as seen from the \textbf{reservoir} module below. 

	With Eq. 3 and A1-A3, the LUTs for each node can be explicitly calculated as 01111111, 0100000000, and 01001101 for nodes 1-3, respectively. The matrix $\bm{\tau}$ specifies the delays in integer multiples of $2\tau_{inv}$. A network with this specification is realized by the module \textbf{reservoir} in Fig. 13 and the \textbf{node} and \textbf{delay\_in} modules described in this section. 

	Like the other modules, \textbf{reservoir} requires a module declaration, parameter declarations, and input/output declarations. Here, we also declare a wire \textbf{x\_tau} that is the delayed reservoir state. In lines 9-11, the nodes are declared with the appropriate parameters and connections and are named \textbf{node\_0}, \textbf{node\_1}, and \textbf{node\_2} respectively. The 6 delay lines are declared and named in lines 13-18. 

\begin{figure}[h]
\centering
\begin{subfigure}[b]{0.45\textwidth}
\captionsetup{font=sf}
\centering
\includegraphics[width=\textwidth]{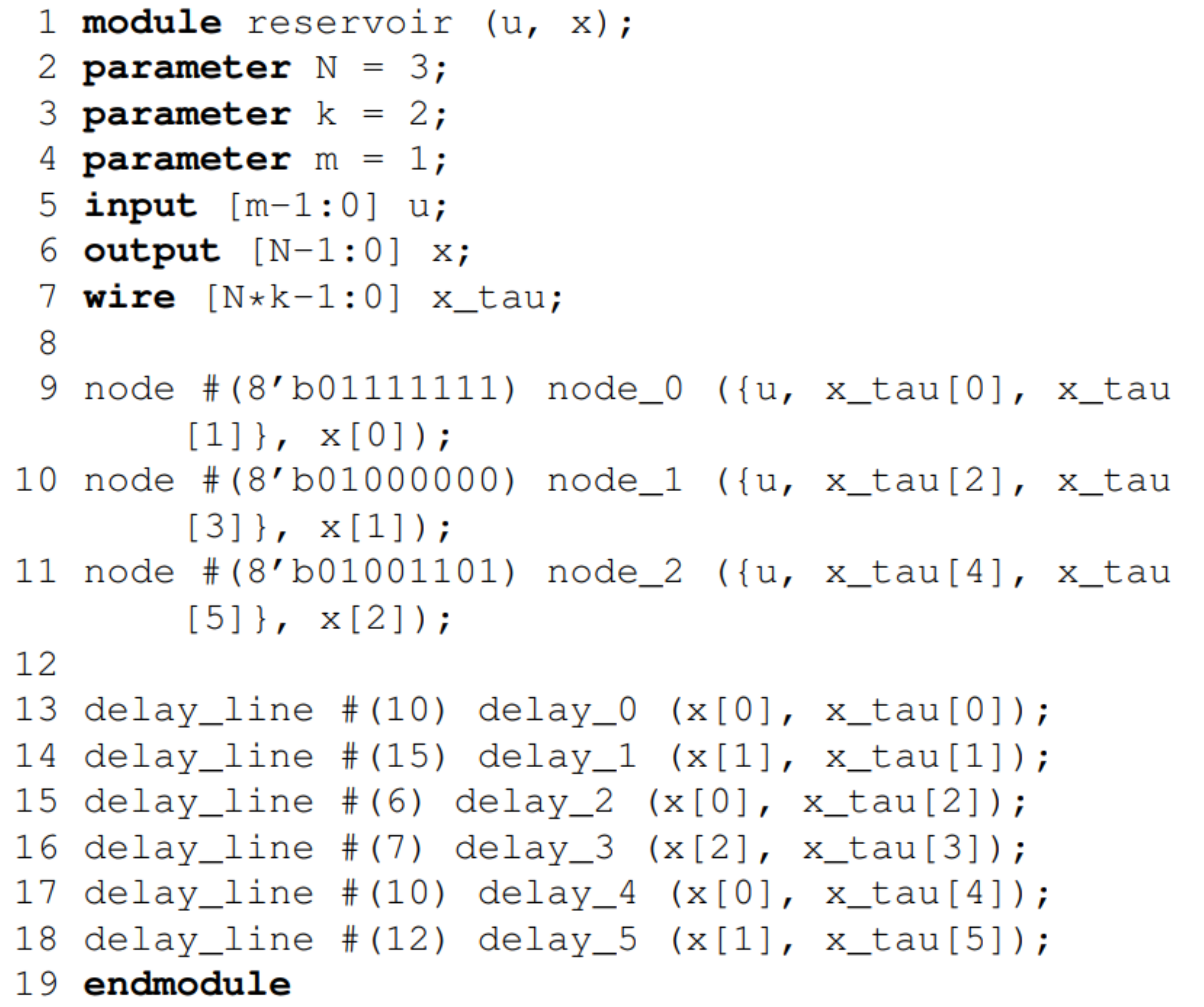}
\end{subfigure}
\caption{Verilog code describing a simple reservoir. The connections and LUTs are determined from Eq. 3 and A1-A3. Lines 9-11 declare 3 nodes. Lines 13-18 declare delay lines that connect them.}
\end{figure}

\section{Synchronous Components}

	In this appendix we discuss the details of the synchronous components that interact with the autonomous reservoir. These components regulate the reservoir input signal, the operation mode (training or autonomous), the calculation of the output signal, and record the reservoir state.
    
    Crucial to successful operation is access to a \textbf{sampler} module that reads data from the reservoir and a \textbf{player} module that writes data into the reservoir. The details of these modules are not discussed here as they depend on the device and the application of the reservoir computer. We assume that these modules are synchronized by a global clock \textbf{clk} such that \textbf{sampler} (\textbf{player}) reads (writes) data on the rising edge of \textbf{clk},

	In Fig. 14 we present a sample Verilog code for a high-level module \textbf{reservoir\_computer} containing the reservoir and synchronous components. An instance of a \textbf{sampler} module is coupled to a global clock \textbf{clk} and outputs an $m$-bit wide signal \textbf{u}, a $1$ bit signal \textbf{mode} that determines the mode of operation for the reservoir, and a $2m(N+1)$-bit wide signal \textbf{W\_out} that determines the output weight matrix. An instance of a \textbf{player} module is also coupled to a global clock \textbf{clk} and inputs an $N$-bit wide signal \textbf{x} and a $m$-bit wide signal $v$. Depending on how these modules are implemented, they may also be coupled to other components, such as on-board memory or other FPGA clocks.
    
\begin{figure}[h]
\centering
\begin{subfigure}[b]{0.45\textwidth}
\captionsetup{font=sf}
\centering
\includegraphics[width=\textwidth]{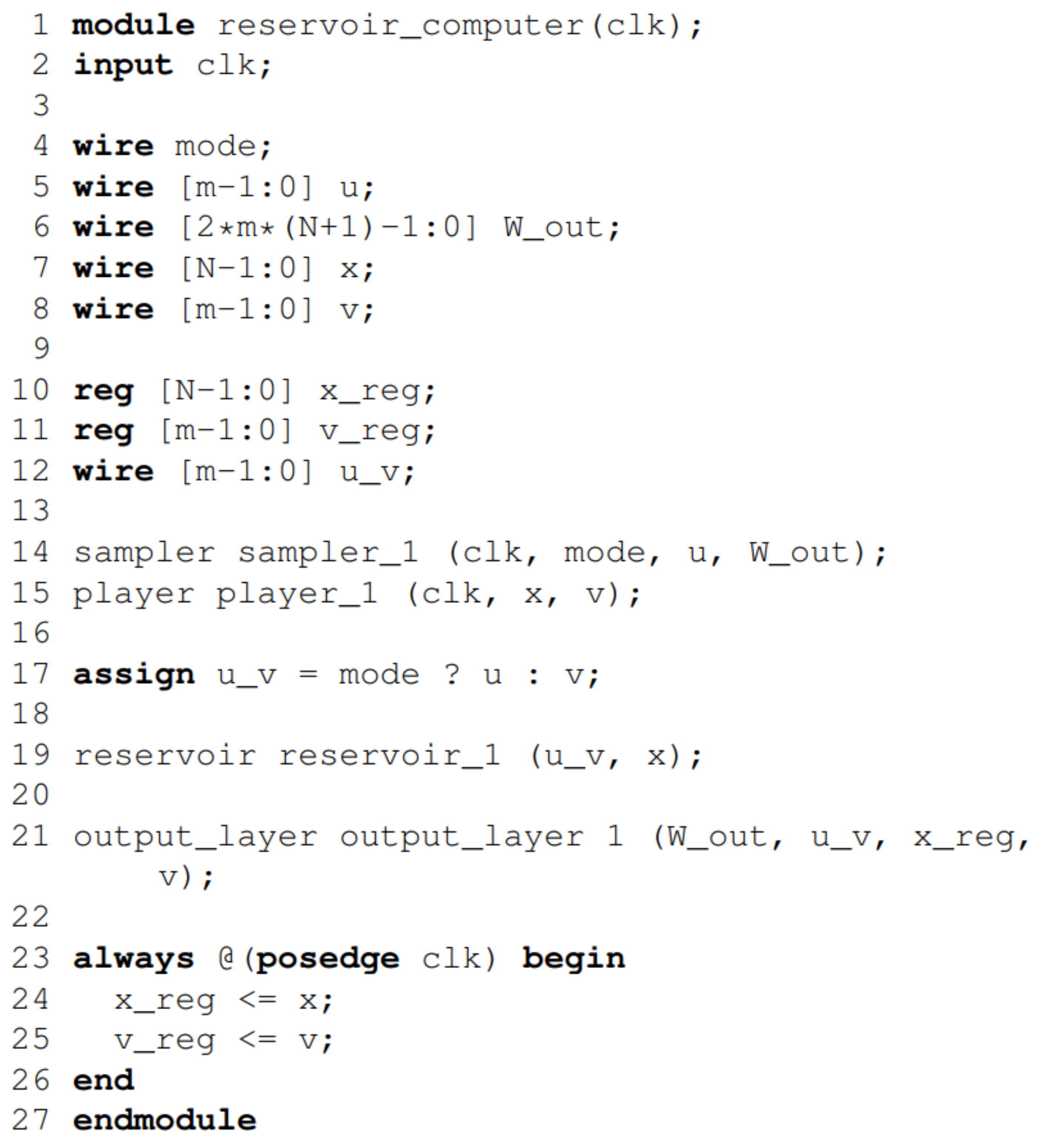}
\end{subfigure}
\caption{Verilog code describing the reservoir computer. It contains the \textbf{reservoir} module discussed in App. A and various synchronous components.}
\end{figure}
    
    As seen in line 17, the state of \textbf{mode} determines whether \textbf{u} or \textbf{v} drives the reservoir. This bit is set to $1$ during training and $0$ after training to allow the reservoir to evolve autonomously (see Fig. 1). 
    
    \textbf{clk} registers \textbf{x} and \textbf{v} so that \textbf{output\_layer} sees a value of \textbf{x} that is constant throughout one period $t_{sample}$ and outputs a value \textbf{v} that is constant over that same interval (see Fig. 3). The module \textbf{output\_layer} performs the operation $\textbf{W}_{out} (\textbf{x}, \textbf{u})$, as described in Sec. V. \textbf{W\_out} is a flattened array of the $N+1$ output weights represented by $2m$ bits, with the extra bits being necessary to avoid errors in the intermediate addition calculations. 
    
\nocite{*}
\bibliography{aipsamp}

\end{document}